\pdfoutput=1
\documentclass[9pt,technote]{IEEEtran}

\IEEEoverridecommandlockouts                              

\usepackage{epsfig} 
\usepackage{amsmath} 
\usepackage{amssymb}  
\usepackage{mathtools}
\usepackage{graphicx}
\usepackage{booktabs,array}
\usepackage{rotating}
\usepackage{caption}
\usepackage{subcaption}
\usepackage{color}
\usepackage[table]{xcolor}
\usepackage{multirow}
\usepackage{tikz}
\usepackage[percent]{overpic}
\usetikzlibrary{arrows,chains,matrix,positioning,scopes}

\makeatletter
\tikzset{join/.code=\tikzset{after node path={%
\ifx\tikzchainprevious\pgfutil@empty\else(\tikzchainprevious)%
edge[every join, bend right]#1(\tikzchaincurrent)\fi}}}
\makeatother
\tikzset{>=stealth',every on chain/.append style={join},
         every join/.style={->}}
\tikzstyle{labeled}=[execute at begin node=$\scriptstyle,
   execute at end node=$]

\newcommand{\indep}{\mathrel{\perp\mspace{-10mu}\perp}} 

\usepackage{tikz}
\newcommand\copyrighttext{%
\footnotesize Submitted to IEEE Transactions on Robotics, October 2017. \copyright 2017 IEEE. Personal use of this material is permitted. Permission from IEEE must be obtained for all other uses, in any current or future media, including reprinting/republishing this material for advertising or promotional purposes, creating new collective works, for resale or redistribution to servers or lists, or reuse of any copyrighted component of this work in other works.}
\newcommand\copyrightnotice{%
\begin{tikzpicture}[remember picture,overlay]
\node[anchor=south,yshift=7pt] at (current page.south) {\fbox{\parbox{\dimexpr\textwidth-\fboxsep-\fboxrule\relax}{\copyrighttext}}};
\end{tikzpicture}%
}

\title{\LARGE \bf Detection and Tracking of General Movable
Objects in Large 3D Maps}

\author{\IEEEauthorblockN{Nils Bore, Johan Ekekrantz, Patric Jensfelt and John Folkesson}
\IEEEauthorblockA{Robotics, Perception and Learning Lab\\
Royal Institute of Technology (KTH)\\
Stockholm, SE-100 44, Sweden\\
Email: \{nbore, ekz, patric, johnf\}@kth.se}}

\begin{document}

\maketitle
\copyrightnotice
\vspace{-10pt} 

\begin{abstract}
This paper studies the problem of detection and tracking of general
objects with long-term dynamics, observed by a mobile robot moving in
a large environment.
A key problem is that due to the environment scale,
it can only observe a subset of the objects at any given time.
Since some time passes between observations of objects in different
places, the objects might be moved when the robot is not there.
We propose a model for this movement in which the objects typically
only move locally, but with some small probability they
jump longer distances, through what we call global motion.
For filtering, we decompose the posterior over
local and global movements into two linked processes.
The posterior over the global movements and measurement associations is sampled,
while we track the local movement analytically using Kalman filters.
This novel filter is evaluated on point cloud data gathered 
autonomously by a mobile robot over an extended period of time.
We show that tracking jumping objects is feasible, and that
the proposed probabilistic treatment outperforms previous
methods when applied to real world data.
The key to efficient probabilistic tracking in this scenario
is focused sampling of the object posteriors.
\end{abstract}
\begin{IEEEkeywords}
Mobile robot, multi-target tracking, movable objects, mapping.
\end{IEEEkeywords}

\IEEEpeerreviewmaketitle

\section{Introduction}

Mobile robots often operate in large environments that can
not be fully observed without moving.
In fact, the floor plan of most buildings is divided into rooms,
which the robot can only visit one at a time.
This setup creates an interesting problem if the robot
is to keep track of specific objects. For example, imagine
a care robot that helps elderly keep track of their belongings, as 
suggested in \cite{koch2007indoor}. Since the robot cannot
be in all places simultaneously, it will not observe most
objects as they move, as is required by classical tracking techniques.
We might instead rely fully on appearance to identify the objects, but this
also fails if the objects are visually similar.
To enable tracking in this scenario, we need to take into account
that the objects might have moved while we were not observing
them. Imagine we are looking for a patient's cell phone, and
that we last saw it in her room. In general, we should expect to
find the object in the same place, or somewhere nearby.
Further, if the robot looks, and the phone is not there, the next
hypothesis is that it should be somewhere within the closed
environment of the patient's department or the care home. Any further
guesses need to be based on knowledge of who might have taken the phone and to where.
These common sense intuitions should form the basis for any
tracking algorithm that works even when the robot does not
continuously observe the targets.

\begin{figure*}[thpb]
	\centering 
	\begin{subfigure}[t]{0.3\textwidth}
 	\centering
 	\includegraphics[scale=0.26]{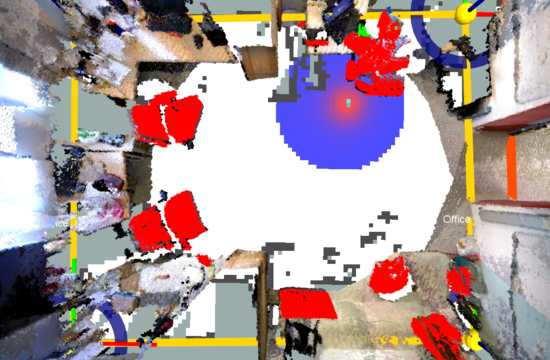}
	\caption{Position probabilities of chair 1.}
 	\end{subfigure}
 	\begin{subfigure}[t]{0.3\textwidth}
  	\centering
  	\includegraphics[scale=0.26]{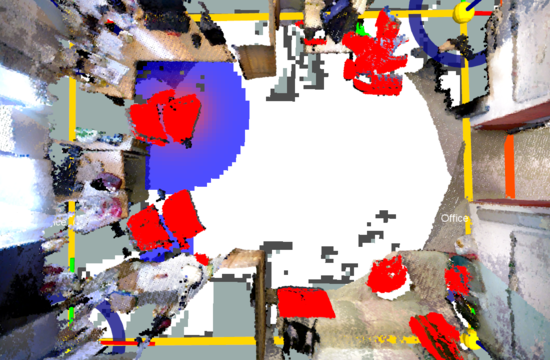}
 	\caption{Position probabilities of chair 2.}
 	\end{subfigure}
 	\begin{subfigure}[t]{0.3\textwidth}
  	\centering
  	\includegraphics[scale=0.26]{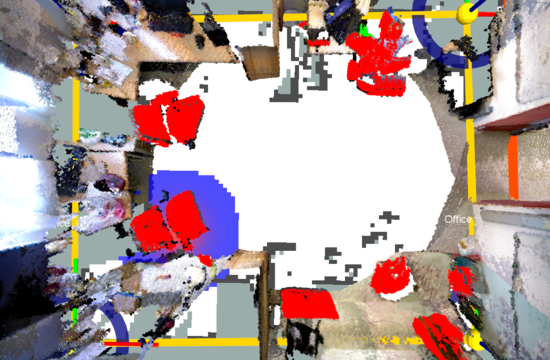}
  	\caption{Position probabilities of chair 3.}
  	\end{subfigure}
   	\caption{Example of our system tracking objects: the current detections of objects (red) and location 		
   			probabilities (blue-violet) for three different chairs. The chairs are
   			visually similar, making long-term tracking just from visual or shape features difficult.}
   	\label{fig:object_maps}
\end{figure*}

The need for tracking specific instances of objects
arises in several robotic applications,
for example when mapping in highly dynamic \cite{wang2007simultaneous} 
and long-term scenarios \cite{biswas2002towards},
obstacle avoidance \cite{du2012robot}\cite{montesano2005modeling} and object search \cite{shubina2010visual}.
Of particular interest to us is surveillance of
objects with long-term dynamics moving in large environments.
The application we have in mind is a robot tasked with
monitoring a number of mostly static objects that typically do not leave some 
closed environment, such as in the care scenario described above.
Another application is security, where a robot should 
monitor the presence for example of important items in an office.
In these surveillance applications, a probabilistic model for the objects' positions
is essential for the robot, since it aids it in knowing where to look.
In addition to helping us distinguish similar objects, a tracking
framework with a probabilistic motion model provides us with
such distributions, as illustrated in Figure \ref{fig:object_maps}.
When the robot covers more locations, the model
should refine the distributions to aid the search.

As mentioned in \cite{gallagher2009gatmo}, tracking of instances is also
necessary when we would like to learn explicit movement models for objects. This is true
both for instance level models and more general object categories.
For example, to estimate a movement model for the mug category, it does not suffice
to study the collective movement of mugs.
This can be illustrated by a collection of mugs standing in a kitchen drying rack.
If the method can only estimate the number of mugs in the rack, it will fail
to detect that people are leaving washed mugs, in order to pick them up later.
Therefore, a reasonable algorithmic model
would be that of a queue, rather than a set.
More generally, a cumulative model will fail to 
capture any such interactions of objects.

Since a mobile robot cannot observe all objects in a
large environment simultaneously, some of them will be moved
when the robot is not looking. Most often, objects only move
locally, as humans might use them for some task and then put
them down. However, objects sometimes move
in unpredictable ways, to entirely different parts
of the robot environment. This might happen if the task is
completed somewhere else than the original object position,
as when drinking from a cup brought from the kitchen.
A key insight is that these latter movements happen much more
seldom than small adjustments of object positions.
We therefore propose decomposing the modeling of object
movement into two parts:
\textit{local movements} describing the small adjustments, and
rarer \textit{global movements} for longer jumps.
In this scheme, we can explain most object detections with local 
movements from the previous positions.
One concrete advantage of this can be highlighted by an example
where we have observed two visually similar objects.
With new observations, the first object
is well explained by local motion from the previous position.
If the second is not but a similar object is observed elsewhere,
we can conclude that the second is more likely to have jumped
there than the first.
This matches well with our human intuition.

To track objects within the maps, we need
to generate object detections from the scene observations.
In previous work on similar topics \cite{gallagher2009gatmo}\cite{biswas2002towards},
it has been popular to use scene differencing for detection.
This is natural, as we are only interested in tracking
objects that move; static objects are trivial to track.
But even movable objects often remain static for long periods
of time, necessitating a mechanism for detection when they
do not move. Methods have been proposed
for extending 3D scene differencing temporally, to segment
one movable object across a sequence of scene observations \cite{herbst2011rgb}.
In this work, we propose restricting such an extension to
only the unambiguous parts. In practice, this means the we
identify two parts in different scenes as the same only
if they are static between the observations. As we will show, such a scheme
can still segment the movable objects in all frames.
Importantly, it defers movable object clustering
to the tracker, which incorporates a probabilistic motion model and
can therefore handle uncertainty.

In our scenario, the task is assumed to be to track
and monitor a fixed set of objects.
The robot patrols one floor of a building at regular intervals,
visits all relevant locations and builds local 3D maps.
From each map observation, we can extract a set of object detections
and corresponding visual features.
In the experiments we use CNN features, which
proved to discriminate well between dissimilar objects.
Although it only takes a few
minutes to travel between the different locations and construct the maps,
we expect the robot to visit the same location again on the
time scale of a few hours. Since it can only survey one
location at a time, the majority of the objects we
are interested in will not be observed for
hours at a time. Between observations
the objects move locally within the locations and will occasionally
jump to different locations.
The sparse observations and the two movement modes complicate
inferring the object positions and demand careful modeling
of probabilities.

The main idea presented in the paper brings together
modeling of discrete object
jumps in between locations and smaller local variation in a 
probabilistic multi-target tracking framework.
We survey previous work of relevance
and present a key set of assumptions which make the problem tractable.
Since the method is developed for tracking in the context of a mobile
robot patrolling an environment, we have developed a complete
system to verify the viability of the approach.
In summary, we present the following contributions:
\begin{enumerate}
\item A multi-object tracking model for continuous movement paired with
      larger discontinuous movement that happens when the robot is somewhere else
\item A practical, efficient inference scheme that fully captures the proposed underlying model
\item Experimental validation on a robot and in simulation
\end{enumerate}

\section{Related Work}

This paper sits atop and draws upon a variety of different
disciplines, including mapping, computer vision and Bayesian inference.
As such, we focus on a few key areas where we find
the most relevant work.
Most studies that address detection and tracking of
movable objects in a mobile robotic framework are found within
the area of \textit{robot mapping}.
In this field, considerable effort has been spent to handle dynamics
in order to better model noise in the measurements and to improve
localization \cite{wolf2005mobile}\cite{biber2005dynamic}.
Others aim to infer object positions with respect to the map,
as described in the sections below.
Our work is a variant of the
\textit{multi-target tracking} problem; another area of relevance which
we discuss in the subsequent paragraphs.
Finally, we survey related work on \textit{3D object detection and segmentation}.

Within the fields of \textit{robot mapping and localization}, 
there have been substantial contributions to solving
detection and tracking of multiple objects,
referred to as the \textit{DATMO} problem \cite{wang2002simultaneous}.
The area can be divided into approaches dealing with
short-term dynamics such as cars and humans \cite{wang2007simultaneous}\cite{montemerlo2002conditional},
and long-term dynamics, such as furniture \cite{schulz2001probabilistic}\cite{gallagher2009gatmo}\cite{biswas2002towards}\cite{anguelov2002learning}.
In a robotic setting, short-term dynamic objects are typically
continuously observed while long-term, or \textit{semi-static}, objects are only
observed every once in a while, during which the objects might
have moved.
In the following, we will focus on methods that deal
with the latter problem, corresponding to our problem formulation.
Biswas et al. \cite{biswas2002towards} studies the problem of associating
dynamic objects between measurements at different times.
Besides the static map, they also keep track of a set
of objects, each associated with at set of
measurements.
While \cite{biswas2002towards} do not incorporate a movement model,
thus allowing arbitrary movement, they use the \textit{mutual
exclusion constraint}, which disallows association of two detections
made at the same time to the same object.
A more recent system dealing with objects that are moved
was presented in \cite{toris2017temporal}. The authors specifically
address the problem of modeling how long objects can be expected to
stay at their last observed position. In addition, they present
a scheme for learning the model from observations. While we
do not employ the proposed probabilistic model for this time
interval, it can be integrated into our system by modifying
our prior model.

The \textit{GATMO} system \cite{gallagher2009gatmo} presented
by Gallagher et al. is
the method that comes closest to realizing the vision
we present in this paper.
In fact, that system addresses the problem of tracking
short-term, as well as long-term dynamics.
The method detects movable objects
by means of scene differencing on 2D laser data
The discovered objects are kept in a database of categories,
with an object moving to an
"Absent" category if it is no longer in a similar position as when
last observed. It may move back to the "Movable" category if there is
a new detection somewhere else with a similar laser signature.
Also worth noting here is earlier work of
Schulz et al. \cite{schulz2001probabilistic} and Wolf et al. \cite{wolf2003towards},
who present similar systems, but only for tracking moveable
objects locally within a map.
Notably, \cite{schulz2001probabilistic} also incorporates the
localization uncertainty of the robot when estimating the object positions.
GATMO \cite{gallagher2009gatmo} and to some extent 
\cite{schulz2001probabilistic}\cite{wolf2003towards}
address the same general problem that we do.
We improve on \cite{schulz2001probabilistic}\cite{wolf2003towards}
by also tracking larger scale motion, necessitating
joint modeling of the objects.
Our probabilistic model is an important improvement on
the GATMO \cite{gallagher2009gatmo} system.
As demonstrated by our experiments,
it enables robust performance also in the presence of noise.
Lastly, unlike these earlier systems,
the proposed system operates on top of 3D maps.

At the basic level, our problem is a \textit{multi-target tracking} (MTT) problem, .
Many techniques have been proposed to address MTT but
few address the problem in a setting similar to ours.
Our system must track the discrete process of jumping between
a finite number of locations as well as the continuous 2D position
of the objects.
Typically, this also includes inferring the association of measurements to targets.
However, explicitly representing associations between targets, locations
and measurements requires a high-dimensional state space and is not
feasible. Instead, we focus on sampling based \textit{particle filters}.
To improve performance in high-dimensional state spaces
one can sometimes use analytic representations 
of the continuous part, resulting in a so-called
\textit{Rao-Blackwellized particle filter} (RBPF).
RBPFs have been proposed for multi-target tracking in
a number of papers, e.g. \cite{sarkka2007rao}\cite{oh2009markov}\cite{miller2007rao}.
Our approach is inspired by that of S{\"a}rkk{\"a} et al. \cite{sarkka2007rao}, 
which integrates one measurement at a time and samples correspondences
to the tracked targets. Given the associations, the continuous
positions can be tracked using classical Kalman filters.
However, our sampling scheme is more
similar to the method of Vu et al. \cite{vu2014particle}, which
uses \textit{MCMC} techniques to estimate the associations
of several measurements jointly.
Another approach to MTT is \textit{Probability Hypothesis Density} (PHD) 
filters \cite{mahler2001multitarget}\cite{vo2006gaussian},
which maintain the combined target intensity function rather than the posteriors.
As we are interested in explicit object posteriors, e.g. for object
search, these methods are not directly relevant to this paper.

In the broader field of \textit{3D object detection and segmentation},
there are several works relevant to ours, particularly
in regards to segmentation.
In \cite{herbst2011rgb}, Herbst et al.
present a system which jointly segments scenes observed
at different times. The method uses an elegant
multi-scene MRF formulation and graph cuts to segment objects that move in between observations.
This has the advantage of providing consistent segmentation 
boundaries by comparing several static scenes to each other at 
once, thus filtering out noise.
While the output of this segmentation is the same as ours,
it is unlikely that method would scale from the tens of scenes
presented in the paper to the hundreds that might be collected by a robot.
Instead, our method can run in an online manner,
which is desirable on a robot platform.
In \cite{finman2013toward}, Finman et al. present a complete
system for object discovery.
Their method detects objects from pair-wise
scene differencing and associates new
detections with previously observed object models.
Similar to us, they
update the visual object models online using a tracker, but do not
explicitly model specific object instances or their positions.
In \cite{ambrus2014meta} we introduced a scene differencing method 
that was also used in \cite{ambrus2015unsupervised}.
The method finds objects by comparing to
a background model of the environment, called a \textit{meta-room}.
In the current work, we do not rely on a background model.
Instead, we compare precisely registered depth frames of subsequent observations,
allowing us to reliably detect smaller objects than \cite{ambrus2014meta}, such as mugs.
Another advantage to \cite{ambrus2014meta} is the option of joint segmentation,
allowing us to segment static objects in the past if they move in subsequent observations.

In a subsequent work \cite{ambrus2015unsupervised}, we studied the
related problem of offline clustering of object instances that
have been observed within one room at different times.
One of the main conclusions was that lighting variation or
inability to always observe objects from similar angles present
problems for object recognition in this application.
To alleviate this problem,
we proposed to group the initial clusters of visual features
by spatio-temporal coherency. Importantly, the second
step makes sure no objects observed at the same
time are assigned to the same cluster.
In the current work, we approach the problem from a different
angle, by studying objects that may also move between
locations but never completely leave or enter the robot environment.
Further, while the offline clustering algorithm in
\cite{ambrus2015unsupervised} is greedy, we propose an
online probabilistic framework where we perform joint
data association.

To summarize, while \cite{ambrus2015unsupervised}\cite{wolf2003towards}\cite{schulz2001probabilistic}
track local movement using features and
\cite{biswas2002towards}\cite{anguelov2002learning}
track only using features, ignoring position,
none of them formulate a full motion model that can track general objects.
Only \cite{gallagher2009gatmo} studies
the full problem of tracking general long-term dynamic objects within large environments, but without a specified motion model or a noise model.
We improve on their work by incorporating motion and noise priors in a
probabilistic model working on 3D data. This enables reliable
estimates even in the noisy environments where many robots find themselves.
Detections come from a simple temporally consistent segmentation logic
that combines the precision of \cite{herbst2011rgb}
with the fast sequential updates of \cite{ambrus2014meta}.

\section{Method}

A robot moves between a finite number of locations $l \in \mathcal{L}$ in an environment.
Its task is to monitor a number of semi-static
objects $j \in 1 \dots N$ at each time step $k$.
For this, the robot needs to reason about the current locations $l_{j, k}$ of the objects,
and their exact positions $\mathbf{\hat{x}}_{j, k}$ within the locations.
At each time $k$, it observes one of the locations $l_k^y$, giving
it a sequence of $M_k$ point measurements
$\mathbf{Y}_k = \left\{ \mathbf{y}_{1, k} , \dots , \mathbf{y}_{M_k, k} \right\}$.
Each point comprises a 2D position $\mathbf{\hat{y}}_{m, k}^s$,
together with a visual feature vector $\mathbf{\hat{y}}_{m, k}^f$, each with some noise.
While some of the measurements correspond to one of the $N$ objects
it is monitoring, others originate from other objects or spurious noise.
To know which objects to monitor,
the initial positions $\mathbf{\hat{x}}_{1, 0}, \dots, \mathbf{\hat{x}}_{N, 0}$ are given.
Our tracking formulation is based on the closed world assumption, meaning
no object $j$ enters or fully leaves the environment. This is
justified, since the tracking system can still maintain a distribution over
possible positions in the environment, while concluding it
does not know where an object is if the uncertainty grows too high.

In the following section, we briefly describe how our method generates the point
measurements $\mathbf{Y}_k$ from RGBD scans of the current location $l_k^y$.
Then in subsequent Sections \ref{sec:local_movement} and \ref{sec:global_movement},
we detail the local and global process models, respectively.
Finally, in Section \ref{sec:combined_movement} and onwards,
we describe how to combine the models into a joint posterior and
then propose a framework for efficient inference.

\subsection{Map Differencing and Consistent Segmentation}
\label{sec:map_differencing}

\begin{figure}[t]
\centering
 \includegraphics[width=0.7\linewidth]{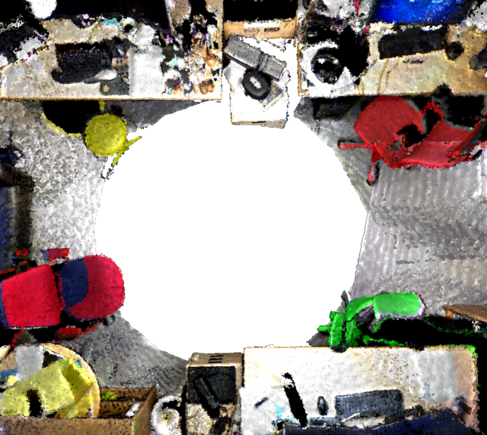} 
  \caption{The change detection result in one local map. There are different sources of
  		   detections, coming from forward change detection (green), backward change detection (red) and
  		   propagated detections (yellow). The hole in the center of the map is due to the
  		   robot standing in that position when collecting the local map.}
  \label{fig:change_detection}
\end{figure}

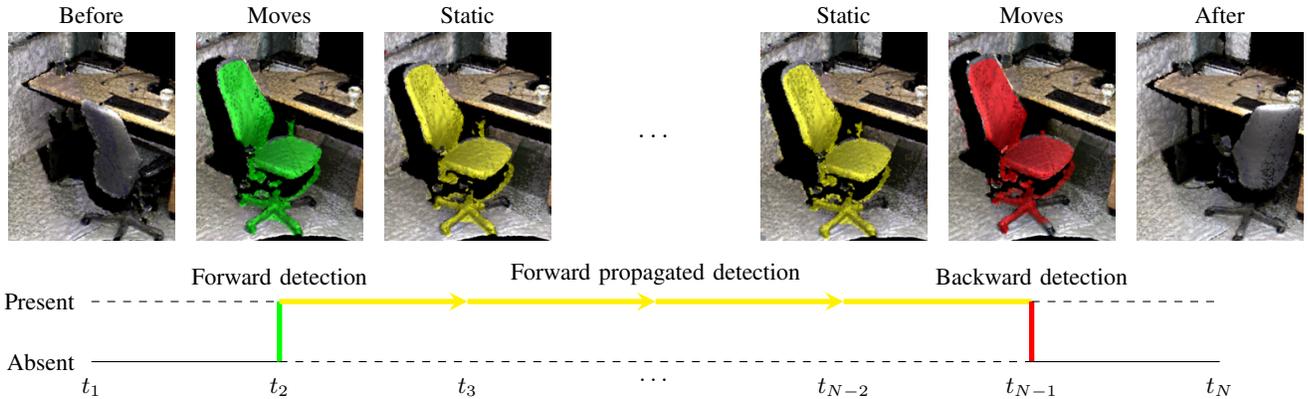
\begin{figure*}[thpb!]
 \begin{center}
  \begin{tikzpicture}[node distance=2.5cm, auto, >=stealth]
   \node[inner sep=0pt, label=above:Before] (a) at (0,0) {\includegraphics[width=0.12\textwidth]{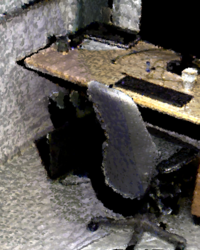}};
   \node[inner sep=0pt, label=above:Moves] (b) [right of=a] {\includegraphics[width=0.12\textwidth]{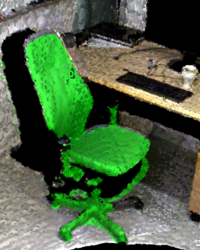}};
   \node[inner sep=0pt, label=above:Static] (c) [right of=b] {\includegraphics[width=0.12\textwidth]{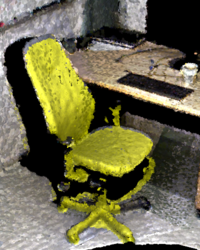}};
   
   \node[inner sep=0pt] (etc) [right of=c] {\large{$\dots$}};
   
   \node[inner sep=0pt, label=above:Static] (d) [right of=etc] {\includegraphics[width=0.12\textwidth]{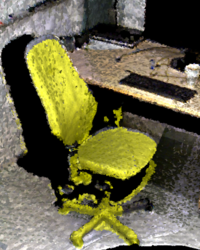}};
   \node[inner sep=0pt, label=above:Moves] (e) [right of=d] {\includegraphics[width=0.12\textwidth]{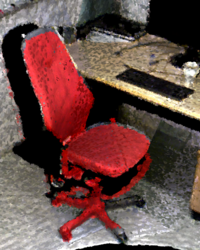}};
   \node[inner sep=0pt, label=above:After] (f) [right of=e] {\includegraphics[width=0.12\textwidth]{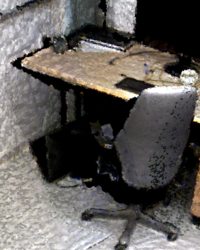}};
   
   \node[below of=a, node distance=2.2cm, label=left:Present] (dummy1) {};
   \node[below of=a, node distance=3.0cm, label=left:Absent, label=below:$t_1$] (graph1) {};
   \node[below of=b, node distance=3.0cm, label=below:$t_2$] (graph2) {};
   \node[below of=b, node distance=2.2cm, label=above:Forward detection] (graph3) {};
   
   \node[below of=c, node distance=2.2cm] (prop1) {};
   \node[below of=c, node distance=3.0cm, label=below:$t_3$] (t3) {};
   \node[below of=etc, node distance=2.2cm, label=above:Forward propagated detection] (prop2) {};
   \node[below of=etc, node distance=3.0cm, label=below:$\dots$] (t4) {};
   \node[below of=d, node distance=2.2cm] (prop3) {};
   \node[below of=d, node distance=3.0cm, label=below:$t_{N-2}$] (t5) {};
   
   \node[below of=f, node distance=2.2cm] (dummy2) {};
   \node[below of=e, node distance=2.2cm, label=above:Backward detection] (graph4) {};
   \node[below of=e, node distance=3.0cm, label=below:$t_{N-1}$] (graph5) {};
   \node[below of=f, node distance=3.0cm, label=below:$t_N$] (graph6) {};
  
   \draw[black] (graph1.center) to (graph2.center);
   \draw[green, line width=0.7mm] (graph2.center) to (graph3.center);
   \draw[yellow, line width=0.7mm, ->] (graph3.center) to (prop1.center);
   \draw[yellow, line width=0.7mm, ->] (prop1.center) to (prop2.center);
   \draw[yellow, line width=0.7mm, ->] (prop2.center) to (prop3.center);
   \draw[yellow, line width=0.7mm] (prop3.center) to (graph4.center);
   \draw[red, line width=0.7mm] (graph4.center) to (graph5.center);
   \draw[black] (graph5.center) to (graph6.center);
   
   \draw[dashed] (dummy1.center) to (graph3.center);
   \draw[dashed] (graph2.center) to (graph5.center);
   \draw[dashed] (graph4.center) to (dummy2.center);
   
  
  \end{tikzpicture}
  \caption{A real world example where a chair moves and then moves again. It is detected through forward change detection and the detection is propagated into subsequent frames. The second movement is detected via the backward change detection pass.}
  \label{fig:prop_illustration}
 \end{center}
\end{figure*}

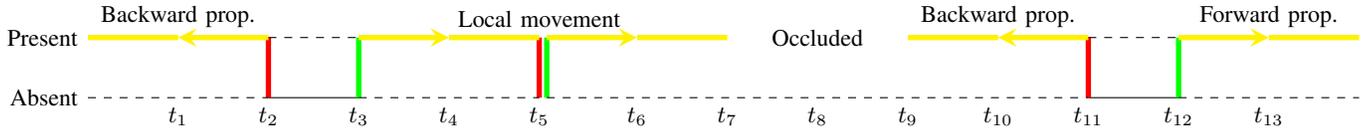
\begin{figure*}[thpb!]
 \begin{center}
  \begin{tikzpicture}[node distance=1.2cm, auto, >=stealth]
   \node[inner sep=0pt, label=left:Present] (p1) at (0,0) {};
   \node[inner sep=0pt, label=above:Backward prop.] (p2) [right of=p1] {};
   \node[inner sep=0pt] (p3) [right of=p2] {};
   \node[inner sep=0pt] (p4) [right of=p3] {};
   \node[inner sep=0pt] (p5) [right of=p4] {};
   \node[inner sep=0pt, label=above:Local movement] (p6) [right of=p5] {};
   \node[inner sep=0pt] (p7) [right of=p6, node distance=0.1cm] {};
   \node[inner sep=0pt] (p8) [right of=p7] {};
   \node[inner sep=0pt] (p9) [right of=p8] {};
   \node[inner sep=0pt, label=center:Occluded] (p10) [right of=p9] {};
   \node[inner sep=0pt] (p11) [right of=p10] {};
   \node[inner sep=0pt, label=above:Backward prop.] (p12) [right of=p11] {};
   \node[inner sep=0pt] (p13) [right of=p12] {};
   \node[inner sep=0pt] (p14) [right of=p13] {};
   \node[inner sep=0pt, label=above:Forward prop.] (p15) [right of=p14] {};
   \node[inner sep=0pt] (p16) [right of=p15] {};
   
   \node[inner sep=0pt, label=left:Absent] (a1) [below of=p1, node distance=0.8cm] {};
   \node[inner sep=0pt, label=below:$t_1$] (a2) [right of=a1] {};
   \node[inner sep=0pt, label=below:$t_2$] (a3) [right of=a2] {};
   \node[inner sep=0pt, label=below:$t_3$] (a4) [right of=a3] {};
   \node[inner sep=0pt, label=below:$t_4$] (a5) [right of=a4] {};
   \node[inner sep=0pt, label=below:$t_5$] (a6) [right of=a5] {};
   \node[inner sep=0pt] (a7) [right of=a6, node distance=0.1cm] {};
   \node[inner sep=0pt, label=below:$t_6$] (a8) [right of=a7] {};
   \node[inner sep=0pt, label=below:$t_7$] (a9) [right of=a8] {};
   \node[inner sep=0pt, label=below:$t_8$] (a10) [right of=a9] {};
   \node[inner sep=0pt, label=below:$t_9$] (a11) [right of=a10] {};
   \node[inner sep=0pt, label=below:$t_{10}$] (a12) [right of=a11] {};
   \node[inner sep=0pt, label=below:$t_{11}$] (a13) [right of=a12] {};
   \node[inner sep=0pt, label=below:$t_{12}$] (a14) [right of=a13] {};
   \node[inner sep=0pt, label=below:$t_{13}$] (a15) [right of=a14] {};
   \node[inner sep=0pt] (a16) [right of=a15] {};
   
   \draw[black] (a3.center) to (a4.center);
   \draw[black] (a13.center) to (a14.center);
   
   \draw[green, line width=0.7mm] (a4.center) to (p4.center);
   \draw[green, line width=0.7mm] (a7.center) to (p7.center);
   \draw[green, line width=0.7mm] (a14.center) to (p14.center);
   
   \draw[red, line width=0.7mm] (a3.center) to (p3.center);
   \draw[red, line width=0.7mm] (a6.center) to (p6.center);
   \draw[red, line width=0.7mm] (a13.center) to (p13.center);
   
   \draw[yellow, line width=0.7mm, ->] (p3.center) to (p2.center);
   \draw[yellow, line width=0.7mm] (p2.center) to (p1.center);
   
   \draw[yellow, line width=0.7mm, ->] (p4.center) to (p5.center);
   \draw[yellow, line width=0.7mm] (p5.center) to (p6.center);
   
   \draw[yellow, line width=0.7mm, ->] (p7.center) to (p8.center);
   \draw[yellow, line width=0.7mm] (p8.center) to (p9.center);
   
   \draw[yellow, line width=0.7mm, ->] (p13.center) to (p12.center);
   \draw[yellow, line width=0.7mm] (p12.center) to (p11.center);
   
   \draw[yellow, line width=0.7mm, ->] (p14.center) to (p15.center);
   \draw[yellow, line width=0.7mm] (p15.center) to (p16.center);
   
   \draw[dashed] (a1.center) to (a3.center);
   \draw[dashed] (a4.center) to (a6.center);
   \draw[dashed] (a7.center) to (a13.center);
   \draw[dashed] (a14.center) to (a16.center);
   
   \draw[dashed] (p3.center) to (p4.center);
   \draw[dashed] (p13.center) to (p14.center);
  
  \end{tikzpicture}
  \caption{A more complex example where the object moves several times and with occlusion.
           Note that small local movements will be registered as a backward and a forward detection
           at the same time step. When occluded, we do not register any measurements.}
  \label{fig:prop_joint_illustration}
 \end{center}
\end{figure*}

As our robot patrols the environment, it performs $360^{\circ}$ 3D
sweeps using its RGBD sensor at the pre-specified locations $\mathcal{L}$.
The RGDB frames are registered into \textit{local maps}, see \cite{ambrus2015unsupervised} for details.
Our aim is to track objects \textit{moving}
within our environment. It is therefore
natural to use \textit{change detection} techniques \cite{ambrus2014meta}\cite{finman2013toward}\cite{herbst2014toward}
to detect image segments corresponding to moving objects.

Change detection only detects objects that move in between two time points $k$ and $k+1$.
However, if at time $k+2$ any of the objects remained in the
same place as in $k+1$, we would like to still detect them,
as this is vital information for our object tracking.
That is, we need to distinguish between not having moved and
not being there any more. Therefore, we add a new
component to the detection system wherein we propagate detections
of objects into previous and subsequent observations.
This allows us to in theory segment all instances of objects
that have moved at any point in the robot's observations.
When formulating the principles for propagating the detected objects, 
there are several scenarios to take into account.
First, an object can appear as well as disappear,
allowing it to be detected by change detection.
We distinguish between detection of appearing and disappearing objects,
by referring to the processes as \textit{forward} and \textit{backward} change detection, respectively.
If an object has appeared in frame $k+1$, it could be present in the 
subsequent frames
while backward detections might have been present in previous frames.
We thus compare all the registered depth image frames in
observation $k+1$ with those in $k+2$ to see if the pixels
corresponding to the object have similar values, see Figure \ref{fig:prop_illustration}.
The object detections are propagated in either direction until the depth
values do no
longer correspond or we detected an object change in the opposite direction,
indicating that the object disappeared, see Figure \ref{fig:prop_joint_illustration}.
Finally, a complicating factor is the objects might be occluded.
While one could use the observations before and after an occlusion to model
the probability of the object being there, we do not to attempt
any such reasoning in this work.

To summarize, several passes determine the final object detections.
First, our algorithm
computes forward- and backward change detection passes by comparing
each subsequent pair of observations. Then, a forward
pass propagates all the detected objects to the subsequent static observations of the same object.
No propagated detection is added when it has already
been detected by another pass. Finally, a backward pass
propagates the detections from backward change detection.
The output of the segmentation of one local map is shown in Figure \ref{fig:change_detection}.
Note that the method can be modified to run online simply by skipping
the backward detection and propagation steps.

\subsection{Object Posteriors}

\begin{figure*}[thpb]
	\centering 
	\begin{subfigure}[t]{0.49\textwidth}
 	\centering
 	\includegraphics[width=0.7\linewidth]{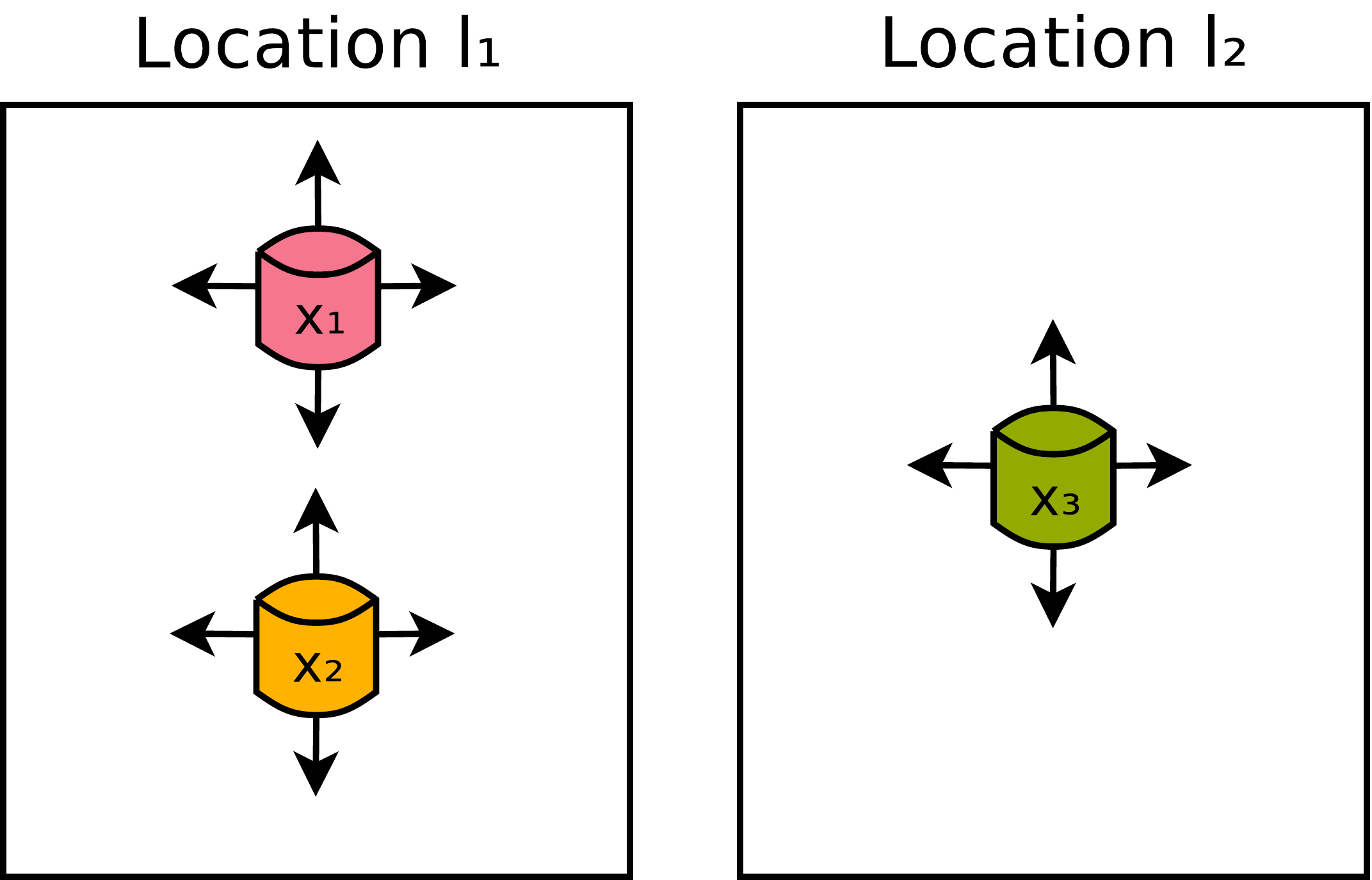} 
	\caption{Local, small movements of the objects within the locations.}
	\label{fig:local_movement}
 	\end{subfigure}
 	\begin{subfigure}[t]{0.49\textwidth}
  	\centering
  	\includegraphics[width=0.7\linewidth]{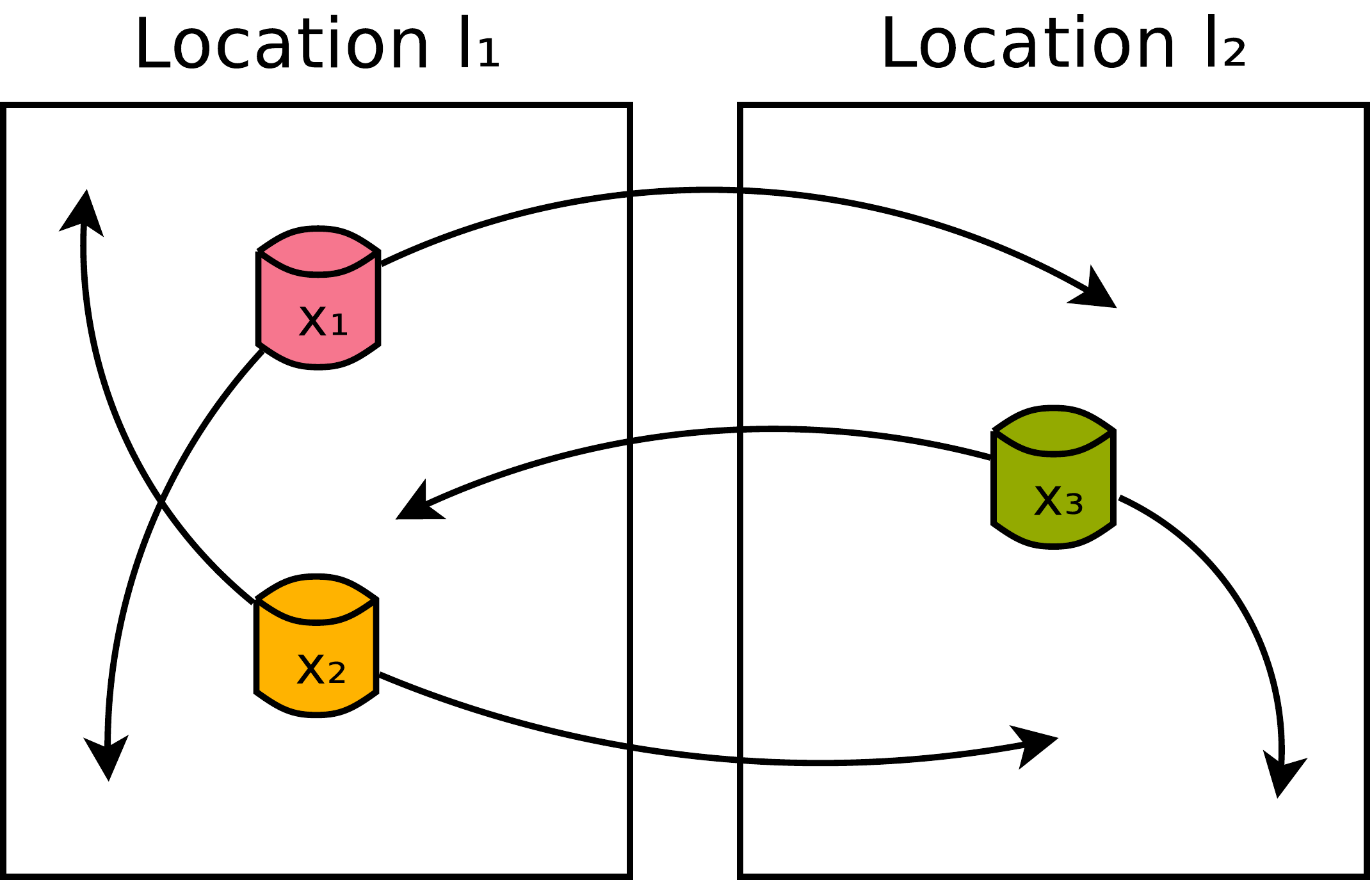} 
 	\caption{Global movements, jumping within or between different locations.}
 	\label{fig:global_movement}
 	\end{subfigure}
   	\caption{Illustration of the possible movements of three objects. Arrows indicate possible movements. The objects can move
   	         locally, within a certain area. Additionally, they may also jump to new
   	         locations in completely different areas of the environment.}
   	\label{fig:movement}
\end{figure*}

Now we turn to the problem of associating the detections from every time 
step with objects in order to track them.
The modeling of object movement can be described in terms of a combination
of two different processes. On the one hand, an object often moves locally
in one place, as illustrated in Figure \ref{fig:local_movement}.
Picture, for example, a chair in front of a desk; it frequently
moves a bit when the user is getting up or adjusting the seating position.
More rarely, objects might also move somewhere completely different, as illustrated in Figure \ref{fig:global_movement}. To be able to track these rare movements, we need
to reason situations when a tracked object's visual
descriptor does not match anything nearby its previous position.
If its descriptor is also close to that of a new, unexplained object, we might be
confident the object has "jumped".
Since the robot can only observe objects within one location $l \in \mathcal{L}$ at each time step,
estimating $l_{j, k}$ for each of the objects
will prove essential to our treatment of the problem.
In the following, we describe our modeling of the two 
processes separately and then how to combine them.

\subsubsection{Local Movement}
\label{sec:local_movement}

Individual object, or \textit{target}, states $\mathbf{x}_{j, k}$
and measurements $\mathbf{y}_{m, k}$ at time $k$ are composed of continuous object 2D position $\mathbf{\hat{\cdot}}^s$ and descriptor $\mathbf{\hat{\cdot}}^f$ vectors, as well as a discrete location $l \in \mathcal{L}$,
$$\mathbf{x}_{j, k} = \left( \begin{bmatrix} \mathbf{\hat{x}}_{j, k}^s \\ \mathbf{\hat{x}}_{j, k}^f \end{bmatrix}, l_{j, k} \right),\:\: \mathbf{y}_{m, k} = \left( \begin{bmatrix} \mathbf{\hat{y}}_{m, k}^s \\ \mathbf{\hat{y}}_{m, k}^f \end{bmatrix}, l_k^y \right).$$
The continuous states $\mathbf{\hat{x}}$ can be observed directly, with some noise and therefore have the
same dimension as the measurements $\mathbf{\hat{y}}$.
Note that for the measurements, the continuous location $\mathbf{\hat{y}}_{m, k}^s$ also uniquely defines the discrete location $l_k^y$.
We denote the set of target states at time step $k$, $\mathbf{X}_k = \left\{ \mathbf{x}_{1, k} , \dots , \mathbf{x}_{N, k} \right\}$ and the measurements $\mathbf{Y}_k$.
If the objects move only locally, and if we assume Gaussian noise
for the process, and for the position and descriptor measurements, we can describe the
system using linear dynamics.
Given assignments $c_{j, k} = m$ that map each target $j$ to its corresponding measurement $m$,
we may use the standard update equations
for the Kalman filter to track each target separately.
With the prior that the object stays locally, the state will
simply propagate as the previous state plus some normally distributed noise.
The continuous states can further be directly observed, giving us
\begin{equation}
\begin{aligned}
\mathbf{\hat{x}}_{j, k} &= \mathbf{\hat{x}}_{j, k-1} + \mathbf{q}_{j, k-1} \\
\mathbf{\hat{y}}_{k, c_{j, k} } &= \mathbf{\hat{x}}_{j, k} + \mathbf{r}_{j, k}
\end{aligned},
\label{eq:linear}
\end{equation}
with $\mathbf{q}_{j, k-1}$ and $\mathbf{r}_{j, k}$ denoting process noise
and measurement noise, respectively. The feature part has no process noise 
but we do assume normally distributed measurement noise. In our
application this is a reasonable assumption, as discussed in Section
\ref{sec:features}. Note that our model can incorporate other
distributions.
In this local model, the location $l_{j, k}$ never changes but it is implicitly given
by the last measurement associated with the target.

\subsubsection{Global Movement \& Associations}
\label{sec:global_movement}

\definecolor{lightgray}{gray}{0.9}

\begin{table*}[htpb]
\begin{center}
\rowcolors{4}{}{lightgray}
\begin{tabular}{r|rrrrrr}
  \multirow{3}{*}{$p(u_{j, k}, l_{j, k}, c_{j, k} | l_{j, k-1})$} & $u_{j, k} = \text{no jump},$ & $u_{j, k} = \text{no jump},$ & $u_{j, k} = \text{jump},$ & $u_{j, k} = \text{jump},$ & $u_{j, k} = \text{jump},$ & $u_{j, k} = \text{jump},$ \\
  & $l_{j, k} = l_{j, k-1},$ & $l_{j, k} = l_{j, k-1},$ & $l_{j, k} = l_k^y,$ & $l_{j, k} = l_k^y,$ & $l_{j, k} = l_\text{unknown},$ & $l_{j, k} = l_\text{unknown},$ \\
  & $c_{j,k} = m$ & $c_{j,k} = \epsilon$ & $c_{j,k} = m$ & $c_{j,k} = \epsilon$ & $c_{j,k} = m$ & $c_{j,k} = \epsilon$ \\
  \hline
  $l_{j, k-1} = l_k^y$ & $\frac{1}{M_k}(1-p_{\text{jump}})p_{\text{meas}}$ & $(1-p_{\text{jump}})(1-p_{\text{meas}})$ & $\frac{1}{M_k N_l}p_{\text{jump}}p_{\text{meas}}$ & $\frac{1}{N_l}p_{\text{jump}}(1-p_{\text{meas}})$ & 0 & $\frac{N_l-1}{N_l}p_{\text{jump}}$ \\
  $l_{j, k-1} \neq l_k^y$ & 0 & $(1-p_{\text{jump}})$ & $\frac{1}{M_k N_l}p_{\text{jump}}p_{\text{meas}}$ & $\frac{1}{N_l}p_{\text{jump}}(1-p_{\text{meas}})$ & 0 & $\frac{N_l-1}{N_l}p_{\text{jump}}$ \\
  $l_{j, k-1} = l_\text{unknown}$ & 0 & 0 & $\frac{1}{M_k N_l}p_{\text{meas}}$ & $\frac{1}{N_l}(1-p_{\text{meas}})$ & 0 & $\frac{N_l-1}{N_l}$ \\
\end{tabular}
\end{center}
\caption{The prior probabilities given previous location. Note that we can get the probability for row two from row one by taking into account that $p(c_{j, k} = m | l_{j, k-1} \neq l_k^y, u_{j, k} = \text{no jump}) = 0$. Similarly, we get the third row from $p(u_{j, k} = \text{jump}) = 1$.}
\label{table:prior}
\end{table*}

In addition to the continuous pose, our objects can also jump between several discrete
locations $l\in \mathcal{L}$. These often correspond to rooms such as offices or
different parts of a hallway.
At each time point, we assume an object might take the "action" $u_{j, k} \in \left\{ \text{jump}, \text{no jump} \right\}$ of
jumping to a random $l$ or staying,
\begin{equation}
p(u_{j, k}) = \left\{
  \begin{array}{ll}
    p_\text{jump}, \text{ if } u_{j, k}=\text{jump}\\
	1 - p_\text{jump}, \text{ if } u_{j, k}=\text{no jump}
  \end{array}.
\right. \\
\end{equation}
If object $j$ did not jump, we know it stayed in the same
location as before, $u_{j, k} = \text{no jump} \Rightarrow l_{j, k} = l_{j, k-1}$.
In the event of a jump, it may uniformly jump to any of the
$N_l = | \mathcal{L} |$ locations in the robot environment. There is therefore the
probability $\frac{1}{N_l}$ of jumping to the location $l_k^y$ we are
currently observing and $\frac{N_l-1}{N_l}$ of jumping to any of the
other locations.
If the number $N_l$ is large, we gain little
information about an object's location from knowing it is
missing from one of the locations. Instead, for simplicity
we introduce a new location $l_{\text{unknown}}$, indicating that
we believe the object jumped to a new location, but we do not know which.
Therefore, if $l_{j, k} = l_{\text{unknown}}$, we already
know $j$ jumped in subsequent time steps, and the location stays
$l_{\text{unknown}}$ until it is associated with a measurement,
$l_{j, k-1} = l_{\text{unknown}} \Rightarrow u_{j, k} = \text{jump}$.
As we will see, this simplification helps improve our inference.
Thus, the location priors are
\begin{equation*}
p(l_{j, k} | u_{j, k} = \text{jump}, l_k^y) = \left\{
  \begin{array}{ll}
    \frac{1}{N_l}, \text{ if } l_{j, k} = l_k^y \\
	\frac{N_l-1}{N_l}, \text{ if } l_{j, k} = l_{\text{unknown}} \\
	0, \text{ for other } l \in \mathcal{L}
  \end{array} .
\right.
\end{equation*}
Intrinsically, multi-target tracking is a problem of estimating the measurement associations $c_{j, k}$.
Now, $c_{j, k}$ can take on measurement indices $m \in 1, \dots, M_k$ as well as $\epsilon$, which indicates
the target did not give rise to any measurement at this time.
If, for example, the last target estimate $l_{j, k-1}$ is not at the
current observation location $l_k^y$, and the target did not jump,
the probability of not getting a measurement of the target is $1$,
since we can only observe one location at a time.
Even if the object is at the estimated location, it is detected with
some large probability $p_\text{meas}$, allowing for some errors in the detector.
Together, this gives us the measurement prior model,
\begin{equation*}
p(c_{j, k} = \epsilon | l_{j, k}, l_k^y) = \left\{
  \begin{array}{ll}
    1, \text{ if } l_{j, k} \neq l_k^y \\
	1 - p_{\text{meas}}, \text{ if } l_{j, k} = l_k^y 
  \end{array},
\right.
\end{equation*}
and correspondingly, if each of the measurements in the location
are a-priori equally likely to originate from the object,
\begin{equation}
p(c_{j, k} = m | l_{j, k}, l_k^y)  = \frac{1}{M_k}\left( 1 - p(c_{j, k} = \epsilon | l_{j, k}, l_k^y) \right).
\end{equation}

With the conditional priors in hand, we can compute
the individual target association priors $p(c_{j, k} | l_k^y, l_{j, k-1}) = p(c_{j, k} | l_k^y, l_{j, k}) p(l_{j, k} | l_k^y, l_{j, k-1}, u_{j, k}) p(u_{j, k})$, see Table \ref{table:prior}.
Since the associations encode the locations $l_{j, k}$ except for when $c_{j, k} = \epsilon$,
we simplify notation by taking $c_k$ to mean all associations and locations at time $k$.
Note that the individual association priors are conditioned only
on $c_{j, k-1}$, as opposed to the full $c_{k-1}$. Further, there is no
closed form expression to combine them into a joint prior
$p(c_k | c_{1:k-1}, \mathbf{Y}_{1:k-1})$, since we need to
disallow assignments of targets to the same measurement.
While we never sample from this specific distribution, it is relevant
to talk about how one might do so here; we will use the same
techniques later on for the proposal distribution.

One option to sample from the joint association prior 
is to use \textit{MCMC} methods.
In our case, one might use blocked
Gibbs sampling \cite{geman1984stochastic} of two random target
assignments at a time, conditioned on the other assignments.
It is important to sample the assignments in blocks of several targets
since this allows the targets to switch measurement assignments.
The Gibbs sampling procedure allows us to keep the probability of assigning a 
measurement in the current room fixed, even when most of the
observations are already assigned to another target.
In effect, this is analogous to adaptively changing the
set of measurements $\mathbf{Y}_k$ depending on which are
unassigned at that iteration of the sampling procedure,
and sample from $p(c_{j, k} | l_k^y, l_{j, k-1})$ computed
over this modified set of measurements.

While we have found that Gibbs sampling of the joint prior is 
indeed feasible, it might be unnecessarily slow.
We can also approximate the distribution using
the assumption that the target locations $l_{j, k}$ are
independent. However, there is still the constraint
over $c_k$ that no two targets can be associated with the same
measurement, giving
\newcommand{\appropto}{\mathrel{\vcenter{
  \offinterlineskip\halign{\hfil$##$\cr
    \propto\cr\noalign{\kern2pt}\sim\cr\noalign{\kern-2pt}}}}}
\begin{equation}
\tilde{p}(c_k | c_{1:k-1}, \mathbf{Y}_{1:k-1}) \appropto \left\{
  \begin{array}{ll}
    0, \text{ if } \exists j \neq j': c_{j, k} = c_{j', k} \neq \epsilon \\
    \prod_j p(c_{j, k} | l_k^y, l_{j, k-1}) \text{ otherwise }
  \end{array} .
\right.
\label{eq:multinomial}
\end{equation}
In particular, this approximation becomes exact when $M_k \gg N$,
since a target can then be assigned to $l_k^y$ without
any noticeable effect on the other priors.
This distribution is easier to work with as we
can use rejection sampling to sample from the individual
priors and reject any sample set with overlapping assignments.

\subsubsection{Combined Process Model}
\label{sec:combined_movement}
Now, we would like to combine the continuous local processes and the
model for the locations and associations $c_{1:k}$ in such a way that we can estimate
the full posterior $p(\mathbf{X}_{1:k}, c_{1:k} | \mathbf{Y}_{1:k})$ jointly.
Fortunately, given $c_k$ and measurements $\mathbf{Y}_k$, the states $\mathbf{x}_{j}$ are independent. This allows us to decompose the
posterior:
\begin{equation}
\begin{gathered}
p(\mathbf{X}_{1:k}, c_{1:k} | \mathbf{Y}_{1:k}) = p(\mathbf{X}_{1:k} | c_{1:k}, \mathbf{Y}_{1:k}) p(c_{1:k} | \mathbf{Y}_{1:k}) \\
= p(c_{1:k} | \mathbf{Y}_{1:k}) \prod_j p(\mathbf{x}_{j, {1:k}} | c_{1:k}, \mathbf{Y}_{1:k}). 
\end{gathered}
\label{eq:full_posterior}
\end{equation}
This is the basic principle underlying the use of Rao-Blackwellized
particle filters for multiple target tracking, see for example \cite{sarkka2007rao}\cite{miller2007rao}.
It allows us to sample a $c_{1:k}^i$ for every particle $i$ while also
maintaining an analytic filter for each state $\mathbf{\hat{x}}_j^i$.
This decomposition reduces sampling variance,
allowing the use of far fewer particles than if we were to track the full
state $( \mathbf{X}_{1:k}, c_{1:k} )$ using particle sampling.

We will outline our modeling of the combined process dynamics before we delve
further into the details of inference.
In the following, we assume the process has propagated according to the discrete
dynamics in the previous section.
If an object $j$ does not jump, it adheres to the dynamics in Equation \ref{eq:linear}. 
If it does jump, the movement model a-priori distributes the spatial part uniformly over the target spatial
domain, $\mathbf{\hat{x}}_{j, k}^s \sim U(\mathcal{X}^s)$.
If at the same time, it is
associated with a measurement $\mathbf{\hat{y}}_{m, k}$,
our new estimate of the target position will therefore be $N(\mathbf{\hat{y}}_{m, k}^s, \mathbf{R}_k^s)$, where $\mathbf{R}_k^s$ is the spatial measurement noise.
The complete continuous dynamics are
\begin{equation}
\begin{aligned}
\mathbf{\hat{x}}_{j, k}^s &= \mathbf{\hat{x}}_{j, k-1}^s + \mathbf{q}_{j, k-1}^s, &&\text{ if } u_{j, k} = \text{no jump} \\
\mathbf{\hat{x}}_{j, k}^s &\sim U(\mathcal{X}^s), &&\text{ if } u_{j, k} = \text{jump} \\
\mathbf{\hat{x}}_{j, k}^f &= \mathbf{\hat{x}}_{j, k-1}^f + \mathbf{q}_{j, k-1}^f \\
\mathbf{\hat{y}}_{m, k} &= \mathbf{\hat{x}}_{j, k} + \mathbf{r}_{j, k}, &&\text{ if } c_{j, k} = m
\end{aligned},
\end{equation}
where the random parts are distributed according to
\begin{equation}
\begin{aligned}
c_k &\sim p(c_k | c_{1:k-1}, \mathbf{Y}_{1:k-1}) \\
\mathbf{q}_{j, k-1} &\sim N(\mathbf{0}, \mathbf{Q}_k) \\
\mathbf{r}_{j, k} &\sim N(\mathbf{0}, \mathbf{R}_k) \\
\end{aligned}.
\end{equation}
\begin{figure*}[thpb!]
	\centering
	\begin{subfigure}[t]{0.48\linewidth}
  	\centering
  	\includegraphics[trim=123 67 100 65,clip,width=0.99\linewidth]{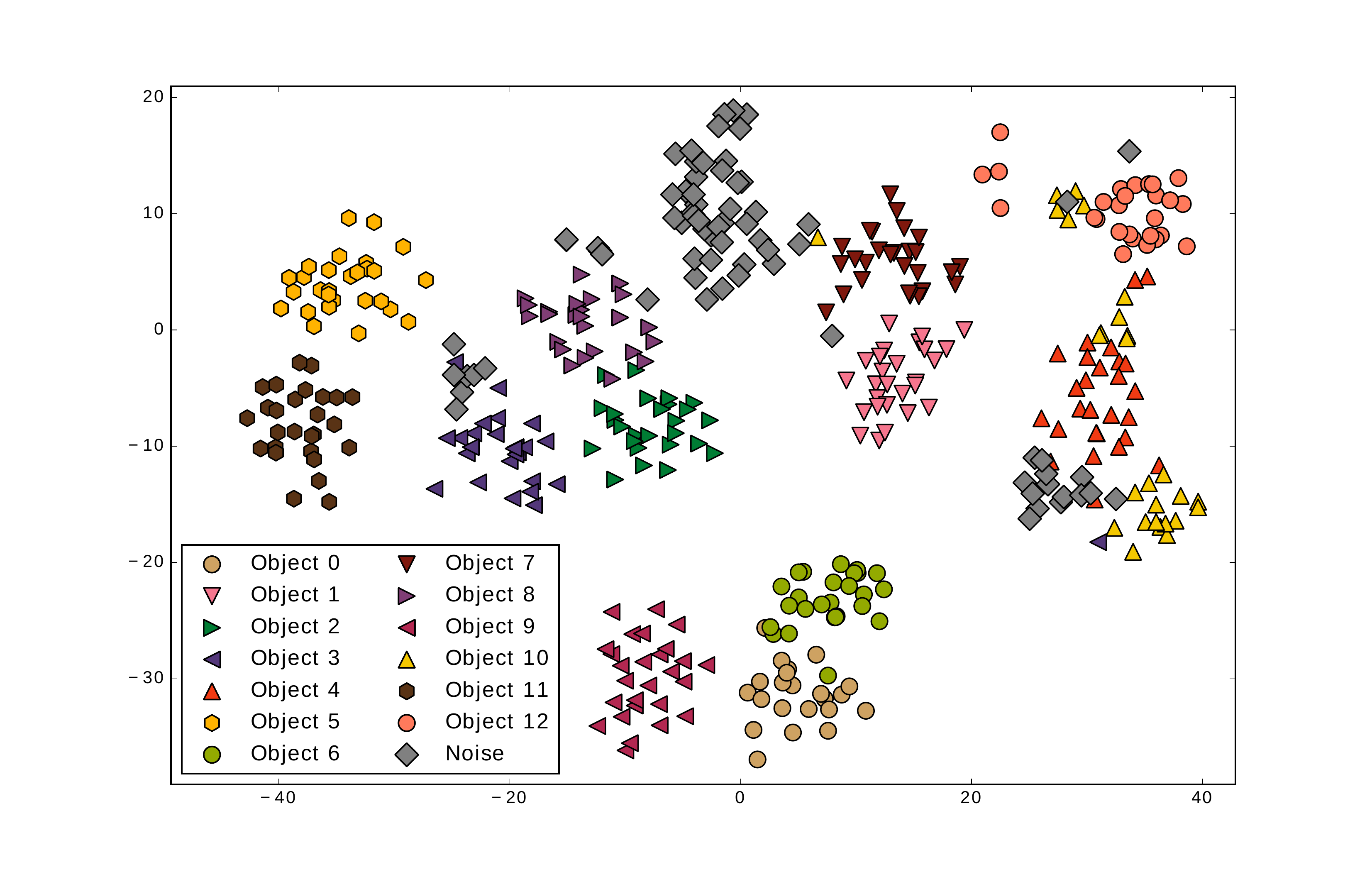} 
    \caption{Features from experiment 1 reduced to two dimensions for 
             visualization. They discriminate well between most of the objects. 
             The only exceptions are the monitor and chairs to the far right: 
             the features of objects 10 and 12 changed drastically after the objects jumped, 
             interleaving them with the noise measurements and those of object 4.}
    \label{fig:experiment2_features}
 	\end{subfigure}
 	\hspace{1em}
	\begin{subfigure}[t]{0.48\linewidth}
 	\centering
 	\includegraphics[trim=119 62 97 60,clip,width=0.99\linewidth]{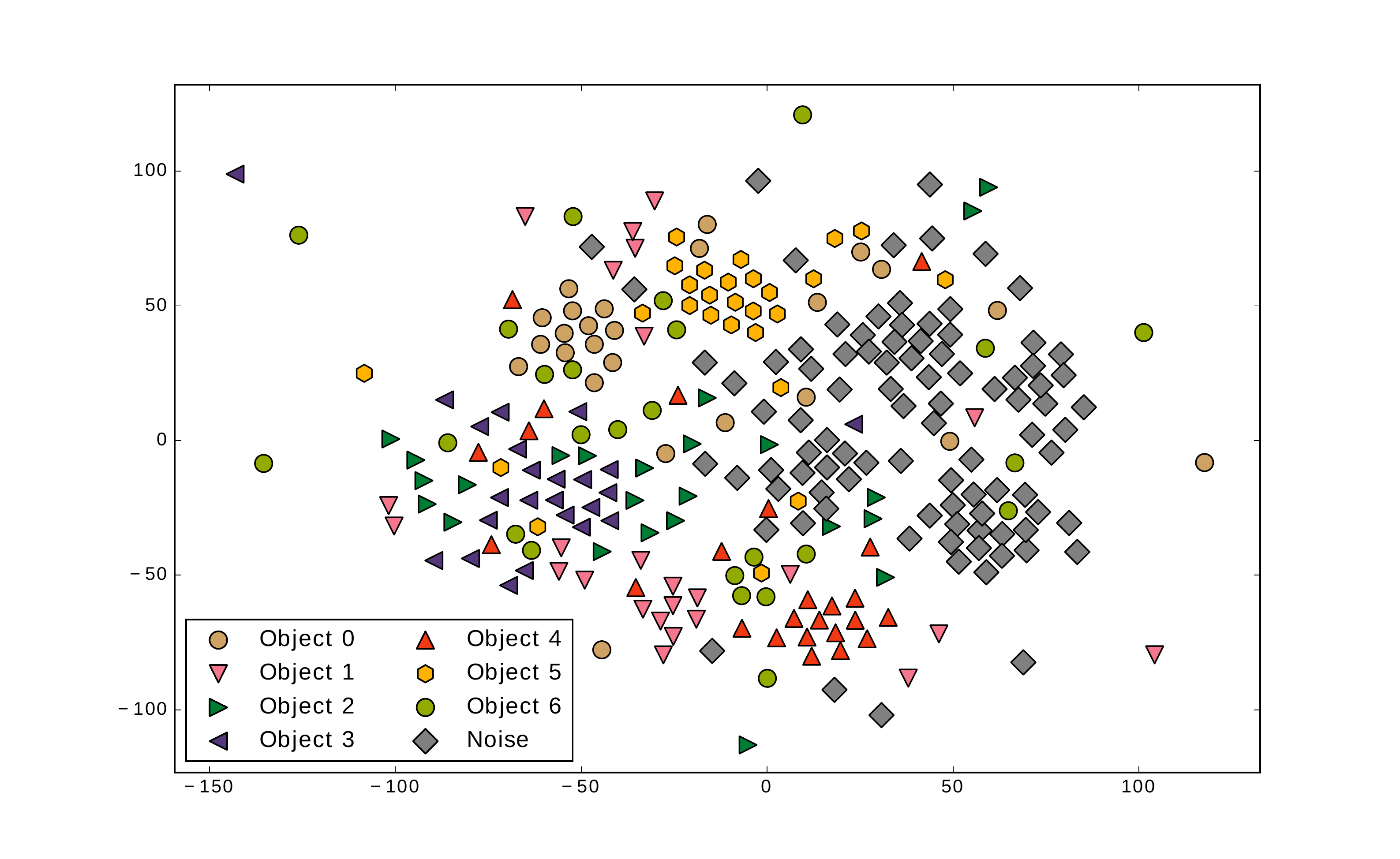} 
    \caption{Features from experiment 2.
             Since the tracked objects are visually similar chairs, the features
             do not discriminate well. However, we see that the noisy detections are
             well separated from the chairs, indicating that they help us
             focus on the objects of interest.}
    \label{fig:experiment1_features}
 	\end{subfigure}
   	\caption{CNN Features from the Google Inception v3 network \cite{szegedy2016rethinking}. They consist of the tensor responses at the final bottleneck layer dimension reduced using t-SNE \cite{maaten2008visualizing}. Additional images and labels for training covariances were extracted from the
  \textit{KTH Longterm Dataset Labels} (\texttt{https://strands.pdc.kth.se/public/KTH\_longterm\_dataset\_labels/readme.html})
  data set. Note that each class corresponds to a specific instance. Semantically similar instances
  are mostly grouped together. For example, the four classes to the left in Figure \ref{fig:experiment2_features} represent all of the food containers, bottles and mugs in that dataset.}
  \label{fig:features2d}
\end{figure*}

\subsubsection{Likelihoods}
\label{sec:likelihoods}

The idea behind our Rao-Blackwellized particle filter is to sample
locations and associations $c_k$ and then update $\left\{ \mathbf{\hat{x}}_j^i \right\}$ using linear Gaussian
dynamics. The idea comes from \cite{sarkka2007rao}, where the system incorporates
one new measurement for every time step. Since our system has distinct time steps where
it gets several coupled measurements at once, we incorporate these
measurements in the same update step, unlike \cite{sarkka2007rao}\cite{miller2007rao}.
In the following, we will look at how to extract weighted samples of the posterior $p(c_{1:k} | \mathbf{Y}_{1:k})$ recursively using a particle filter.
This can then be combined with the linear part of the state to form the full
posterior, see Equation \ref{eq:full_posterior}.
Using Bayes' rule two times, we can decompose the association posterior
to give us a recursive expression:
\begin{equation}
\begin{gathered}
p(c_{1:k} | \mathbf{Y}_{1:k}) \propto p(\mathbf{Y}_{k} | c_{1:k}, \mathbf{Y}_{1:k-1}) p(c_{1:k} | \mathbf{Y}_{1:k-1}) \\
= p(\mathbf{Y}_k | c_{1:k}, \mathbf{Y}_{1:k-1}) p(c_k | c_{1:k-1}, \mathbf{Y}_{1:k-1}) p(c_{1:k-1} | \mathbf{Y}_{1:k-1}).
\end{gathered}
\label{eq:update}
\end{equation}

The first term, $p(\mathbf{Y}_k | c_{1:k}, \mathbf{Y}_{1:k-1}) = \mathcal{L}(\mathbf{Y}_k | c_k)$, defines our 
measurement likelihood.
Given the full set of associations $c_k$, it decomposes into a product of
point likelihoods. Since we know which data points originate from
the tracked objects, we know the others do not. We will refer
to the latter as \textit{clutter measurements}, giving
\begin{equation}
\mathcal{L}(\mathbf{Y}_k | c_k) = \smashoperator{\prod_{j: c_{j, k} = m}} p(y_{m, k} | c_{j, k} = m, u_{j, k}) \smashoperator{\prod_{m: m \text{ is clutter}}} p(y_{m, k} | m \text{ clutter}).
\label{eq:likelihood}
\end{equation}
The point likelihood of local movements
$p(y_{m, k} | c_{j, k} = m, u_{j, k} = \text{no jump})$
is simply the Kalman marginal likelihoods of position and features \cite{sarkka2007rao}.
If we instead consider jumps $u_{j, k} = \text{jump}$, we know 
from the previous section that $\mathbf{\hat{x}}_{j, k}^s$ is 
a-priori uniformly distributed over the spatial domain.
Further, taking the association $c_{j, k} = m$ into account,
it must be somewhere within the measurement location $l_k^y$.
The likelihood is therefore given by the Kalman feature marginal 
likelihood times a uniform density over the area.
With $A_k$ being the area of that location, and
$\boldsymbol{\mu}_{j,k}^f, \boldsymbol{\Sigma}_{j,k}^f$
the feature estimate, we get
$$p(\mathbf{y}_{m, k} | c_{j, k} = m, u_{j, k} = \text{j.}) = \frac{1}{A_k} N(\mathbf{\hat{y}}_{m, k}^f ; \boldsymbol{\mu}_{j,k}^f, \boldsymbol{\Sigma}_{j,k}^f + \mathbf{R}_k^f).$$

As of yet, we have not defined any prior probability of getting
clutter measurements, only of targets not giving rise to a measurement.
However, as we will see, clutter is implicitly sampled
within the Gibbs sampling procedure.
The likelihood of the clutter measurements is given by a uniform
density over the spatial and feature domains, $p(y_{m, k} | m \text{ clutter}) = \frac{1}{A_k S^f}$.
Note that the support of the feature density, $S^f$, needs to
be estimated from data.

\subsubsection{Sampling from the Proposal}
\label{sec:sampling}

A density $q(c_k) \propto p(\mathbf{Y}_k | c_{1:k}, \mathbf{Y}_{1:k-1}) p(c_k | c_{1:k-1}, \mathbf{Y}_{1:k-1})$
proportional to the update of the posterior in Equation \ref{eq:update} is
called the \textit{optimal importance distribution}.
Sampling particle updates from the proposal $q(c_k)$
would minimize variance among 
our particle weights but is in general difficult and typically
requires some approximation \cite{doucet2000sequential}.
Using approximate MCMC methods to sample from the
proposal is therefore a well-established
idea \cite{doucet2000sequential} and has been used to estimate data associations
for multi-target tracking in \cite{oh2009markov}.
In our case, recall that
$p(c_k | c_{1:k-1}, \mathbf{Y}_{1:k-1})$ describes our
transition model, which is in turn given by the individual
association priors as outlined in Section \ref{sec:local_movement}.
The individual priors can be combined with the likelihoods
to compute the individual proposals
$q_j(c_{j, k}) \propto p(\mathbf{Y}_k | c_{j, k})p(c_{j, k} | c_{j:k-1}, l_{j, k-1})$.
Using these distributions, the importance distribution $q(c_k)$ is
sampled analogously to $p(c_k | c_{1:k-1}, \mathbf{Y}_{1:k-1})$;
either with Gibbs sampling or approximated with an independence 
assumption. In the following, we explore the two methods
in more detail.

In Gibbs sampling, the idea is to sample one of the variables $c_{j, k}$
at a time conditioned on all the others, denoted
$c_k^{-j} = \left\{ c_{j', k} \right\}_{j'} \setminus c_{j, k}$.
This gives us a modified individual proposal in the form of a
conditional distribution over $c_{j, k}$,
$q_j(c_{j, k} | c_k^{-j}) \propto p(\mathbf{Y}_k | c_{j, k}, c_k^{-j})p(c_{j, k} | c_{j,k-1}, c_k^{-j})$.
Note that since we know all of the assignments,
$p(\mathbf{Y}_k | c_{j, k}, c_k^{-j})$ can be uniquely
identified with either a target or a clutter likelihood in
Equation \ref{eq:likelihood}.
Again, due to the hard constraints in the prior,
it is important to sample several assignments block-wise.
We do 100 iterations of burn-in,
sampling two random target assignments
from a joint version of $q_j(c_{j, k} | c_k^{-j})$
at each iteration,
and pick the final assignments as our sample.
The algorithm is initialized with assignments 
from approximate rejection sampling as described below.

We also investigate a faster sampling scheme using the
prior of Equation \ref{eq:multinomial}, paired
with approximate independent likelihoods.
The data points are already independent given assignments, $y_{m, k} \indep y_{n, k} | c_{j, k} = m$.
But given no assignment $c_{j, k} = \epsilon$, without the other
assignments we do not know if it implies that a measurement
will be rejected as clutter.
Instead, we approximate it with an independent likelihood 
$p(y_{m, k} | c_{j, k} = \epsilon)$ that should be the same
for all data points, see Section \ref{sec:implementation} for a
derivation of the approximation used here.
Given the likelihood, a joint independent distribution over $c_k$ is given by
$\tilde{q}(c_k) = \prod_j p(y_{m, k} | c_{j, k})p(c_{j, k} | c_{j:k-1}, l_{j, k-1})$. To sample from the approximate posterior,
we generate samples from $\tilde{q}(c_k)$, which we reject if any
two targets are assigned to the same measurement.
In the experiments, we report results both from
this approximate rejection sampling and from Gibbs sampling.

\subsection{Calculating the Weights}
\label{sec:weights}

Since the likelihood $p(\mathbf{Y}_k | c_{1:k}, \mathbf{Y}_{1:k-1})$ is not actually a probability mass
function with respect to $c_k$, i.e. it does not sum to one, the importance distribution is defined by 
the product modulo a normalization constant $Z_k^i$,
\begin{equation*}
q(c_k) = \frac{1}{Z_k^i} \mathcal{L}(\mathbf{Y}_k | c_k) p(c_k | c_{1:k-1}^i, \mathbf{Y}_{1:k-1}),
\end{equation*}
and $Z_k^i$ is given by the
sum of the product over all assignments,
\begin{equation}
Z_k^i = \sum_{c_k} \mathcal{L}(\mathbf{Y}_k | c_k) p(c_k | c_{1:k-1}^i, \mathbf{Y}_{1:k-1}).
\label{eq:weight_update}
\end{equation}
Since $Z_k^i$ varies between the particles, we need
to update the weights proportional to their values of $Z_k^i$,
as it is not reflected in the sampling.
Intuitively, this is similar to a particle filter
that uses the likelihood to update the weights
of the particles.
As we can not compute the $Z_k^i$ directly, it needs to be approximated.
Components of the sum in Equation \ref{eq:weight_update} typically take on their largest
values in a few places where the likelihood is large.
The idea is therefore to produce estimates of $Z_k^i$ by sampling from the proposal
distribution $q(c_k) \propto p(c_k | c_{1:k-1}^i, \mathbf{Y}_{1:k-1}) \mathcal{L}(\mathbf{Y}_k | c_k)$. 
In the following, we mainly use the fact that
$\sum_{c_k} p(c_k | c_{1:k-1}^i, \mathbf{Y}_{1:k-1}) = 1$:
\begin{align}
\begin{split}
Z_k^i &= \frac{Z_k^i}{\sum_{c_k} p(c_k | c_{1:k-1}^i, \mathbf{Y}_{1:k-1})} \\
&= \frac{Z_k^i}{\sum_{c_k} p(c_k | c_{1:k-1}^i, \mathbf{Y}_{1:k-1}) \mathcal{L}(\mathbf{Y}_k | c_k) / \mathcal{L}(\mathbf{Y}_k | c_k)} \\
&= \frac{1}{\sum_{c_k} q(c_k) / \mathcal{L}(\mathbf{Y}_k | c_k)}
= \frac{1}{\mathop{\mathbb{E}}_q (1 / \mathcal{L}(\mathbf{Y}_k | c_k))}.
\end{split}
\end{align}
If we look at the last line, we can then estimate the expectation
$\mathop{\mathbb{E}}_q (1 / \mathcal{L}(\mathbf{Y}_k | c_k))$ by
sampling many $c_k \sim q(c_k)$ and compute
$1 / \mathcal{L}(\mathbf{Y}_k | c_k)$ for each sample.
The inverse of the estimated expectation gives us our estimate
of the sum $Z_k$.
Importantly, we already perform Gibbs sampling from
$q(c_k)$ to sample our particle proposals. By simply running
a few more iterations of the same MCMC chain, we can use this
procedure to estimate $Z_k$. Being able to produce
both of these properties at the same time caters to the
efficiency and simplicity of the algorithm.

In addition to this scheme, we also evaluate a faster version where
we compute $Z_k$ using the approximated independent posterior $\tilde{q}(c_k)$.
Inserting the corresponding product into the sum of Equation \ref{eq:weight_update}
results in a product over the individual sums,
$Z_k^i \approx \prod_j \sum_{c_{j, k}} q_j(c_{j, k}) = \prod_j \sum_{c_{j, k}} p(y_{m, k} | c_{j, k})p(c_{j, k} | c_{j:k-1}, l_{j, k-1})$.

\subsubsection{Rao-Blackwellized Multi-Target Tracking}
\label{sec:tracking}

In summary, this gives us the sequential importance sampling updates:
\begin{enumerate}
\item Sample new locations and associations, either using Gibbs sampling or approximate rejection sampling:
\begin{equation*}
\begin{aligned}
c_k^i &\sim q(c_k | \mathbf{Y}_{1:k}, c_{1:k-1}^i)
\end{aligned}
\end{equation*}
\item Update the position and feature estimates $\boldsymbol{\mu}_{j,k}^i, \boldsymbol{\Sigma}_{j,k}^i$
\item Update the weights using Gibbs from previous section or with approximate $Z_k^i = \prod_j \sum_{c_{j, k}} q_j(c_{j, k})$:
$$w_k^i = w_{k-1}^i Z_k^i, \:\: w_k^i = \frac{w_k^i}{\sum w_k^i}$$
\end{enumerate}

At each step of the filtering, we can then construct a posterior over
the joint feature and spatial distributions of the objects. The
feature part can be marginalized out simply by removing the corresponding
dimensions from the posterior:
\begin{equation}
p(\mathbf{\hat{x}}_j | \mathbf{Y}_{1:k}) = \sum_i w_i N(\mathbf{\hat{x}}_j; \boldsymbol{\mu}_{j,k}^i, \boldsymbol{\Sigma}_{j,k}^i).
\end{equation}
The resulting density is illustrated in the filtering results, see for example
Figure \ref{fig:object_maps} and Table \ref{table:posteriors}.

\subsection{Implementation}
\label{sec:implementation}

While each particle could potentially need its whole history of associations and
measurements to define the probabilities, in reality we can get away with much less.
Each particle is parametrized by a list $\mathcal{P}_k^i = \{ \boldsymbol{\mu}_{j,k}^i, \boldsymbol{\Sigma}_{j,k}^i, l_{j, k}^i \}_{j = 1:N}$ of sets for the targets $j$.
In particular, the target locations $l_{j, k}^i$ fully specify the probability $p(c_k | c_{1:k-1}^i, \mathbf{Y}_{1:k-1})$ of the proposal distribution,
as seen in Table \ref{table:prior}. $\boldsymbol{\mu}_{j,k}^i$ and $\boldsymbol{\Sigma}_{j,k}^i$ parametrize the marginal likelihoods.

For the version with approximate rejection sampling using $\tilde{q}(c_k)$,
we need to estimate the independent likelihoods
$p(y_{m, k} | c_{j, k} = \epsilon)$
conditioned on no observation of $j$.
These roughly describe the data distribution.
One approach would be to try to directly approximate the
unconditional likelihood of the data points, which is hard in itself. 
However, we have found empirically that another approach works better
in our case.
We proceed using the intuition that these values should remain
fairly constant with respect to the current target $j$.
If the current data
is assigned low values by the Kalman likelihoods, those values should
start approaching the data density.
This allows sampling for example of jumps. With this intuition in mind,
we approximate the data likelihood by the expectation of the data likelihood
with respect to the estimated continuous distribution of target $j$, 
$\mathbf{\hat{Y}} \sim N(\boldsymbol{\mu}_{j,k}, 
\boldsymbol{\Sigma}_{j,k} + \mathbf{R}_k)$. In the derivation, we
use the fact that there should be negligible overlap between the 
different target distributions of
one particle and that $\sum_{c_k: c_{j, k} = m} p(c_k) \approx \frac{1}{N}$:
\begin{equation*}
\begin{gathered}
\mathop{\mathbb{E}}_{\mathbf{\hat{Y}}} \left[ p(\mathbf{\hat{Y}})\right]
= \sum_{c_k} \mathop{\mathbb{E}}_{\mathbf{\hat{Y}}} \left[ p(\mathbf{\hat{Y}} | c_k) \right] p(c_k) \approx
\mathop{\mathbb{E}}_{\mathbf{\hat{Y}}} \left[ p(\mathbf{\hat{Y}} | c_{j, k} = m) \right] \times \\
\sum_{c_k: c_{j, k} = m} p(c_k) \approx \frac{1}{N} \mathop{\mathbb{E}}_{\mathbf{\hat{Y}}} \left[ p(\mathbf{\hat{Y}} | c_{j, k} = m) \right] = \frac{(4 \pi)^{-\frac{D}{2}}}{N \sqrt{| \boldsymbol{\Sigma}_{j,k} + \mathbf{R}_k | }}
\end{gathered}
\end{equation*}
While this is an approximation, we have seen empirically that it
performs better than trying to estimate the data likelihood.
Again, we reason this is due to that the marginal
$p(c_{j, k} = \epsilon | \mathbf{Y}_{1:k})$ is determined mainly
by the ratio to the point likelihoods of the measurement associations.

\section{Experiments}

\begin{figure}[thpb]
	\centering
 	\begin{subfigure}[t]{0.66\linewidth}
  	\centering
  	\includegraphics[width=0.99\linewidth]{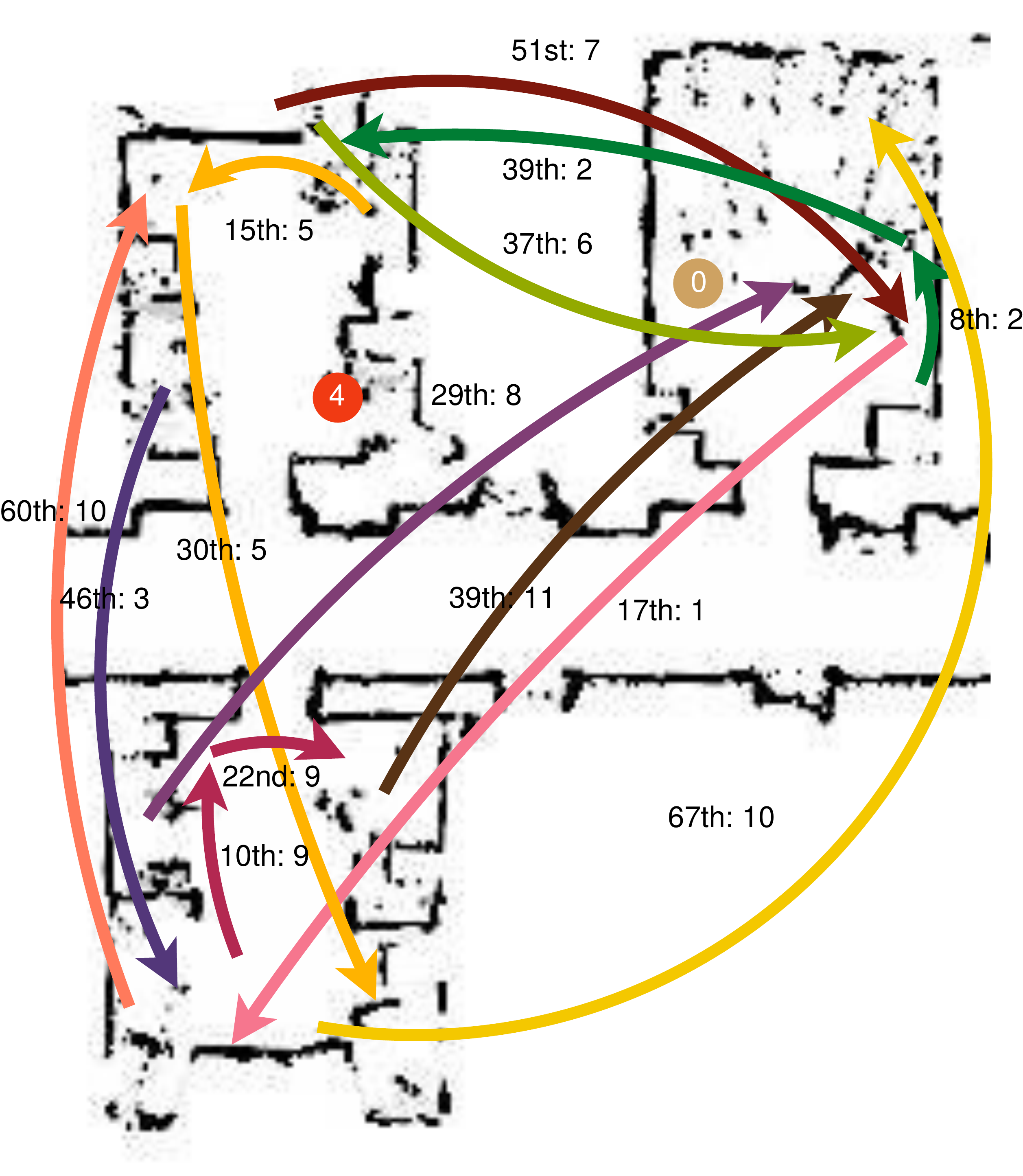} 
 	\caption{Experiment 1.}
 	\label{fig:jumps_experiment2}
 	\end{subfigure}
 	\begin{subfigure}[t]{0.3\linewidth}
 	\centering
 	\includegraphics[width=0.99\linewidth]{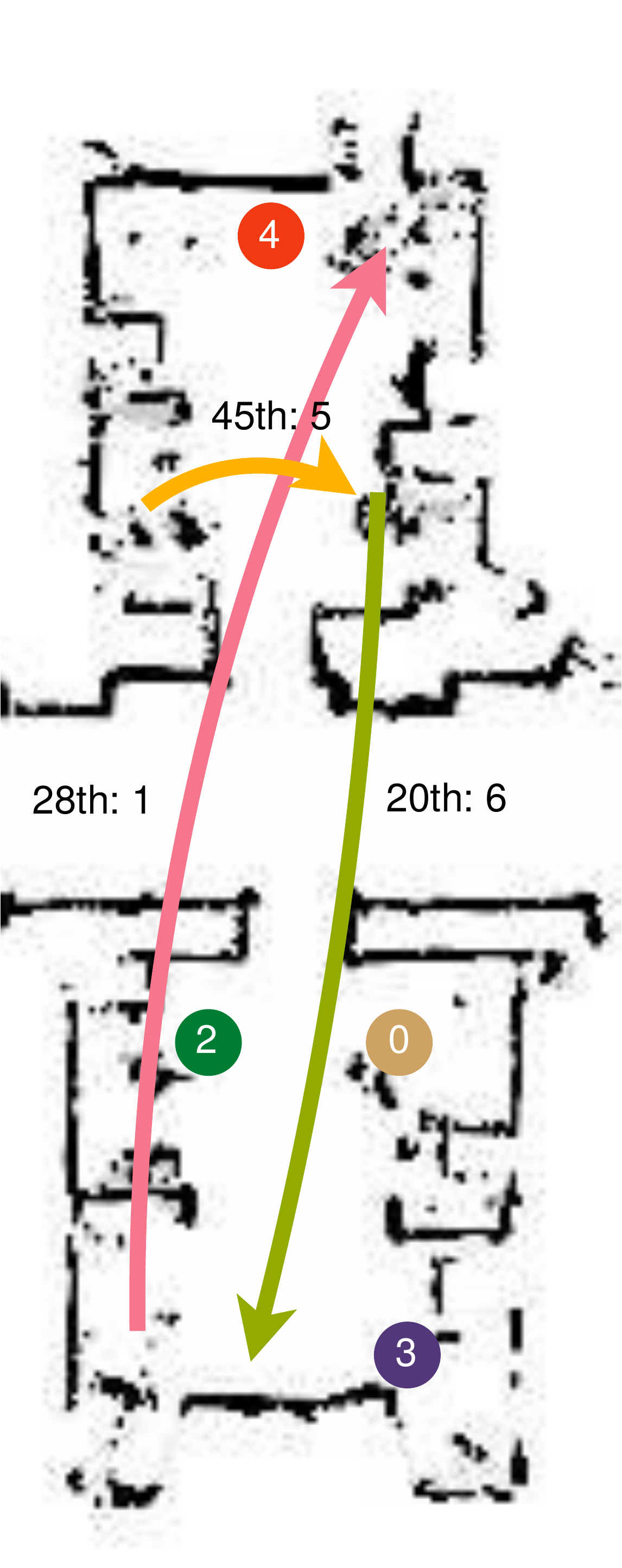} 
	\caption{Experiment 2.}
	\label{fig:jumps_experiment1}
 	\end{subfigure}
   	\caption{The jumps of the objects in the two experiments. Arrows indicate jumping objects while dots show the position of static objects with only local movement. Jumps are annotated with time step and object id. In the first experiment there are 3 jumps between 2 rooms (i.e. locations). The second experiment contains 13 objects from 3 rooms jumping a total of 14 times.}
   	\label{fig:jumps}
\end{figure}

We perform several experiments, both on real data
collected autonomously by a robot and on simulated data.
In the following, we describe the exact details of the
method used as well as the experimental setups.

\subsection{Detections and Features}
\label{sec:features}

\begin{figure*}[thpb]
\centering 
 \includegraphics[width=0.9\linewidth]{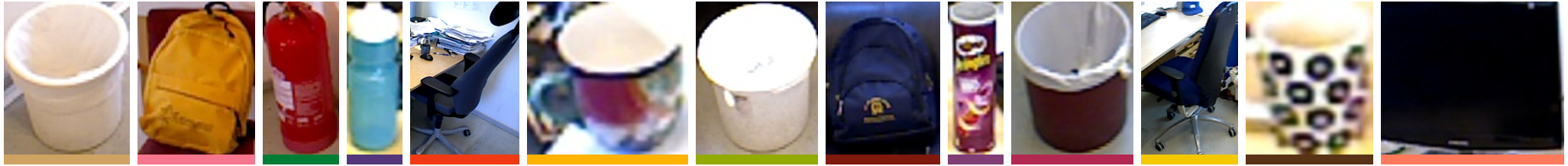} 
  \caption{The objects in experiment 2. Objects 0-12 from left to right: Trash can, bag, fire extinguisher, bottle, chair, mug, trash can, bag, container, trash can, chair, mug, monitor. The tracker features are initialized from these images, which were also extracted using the method in Section \ref{sec:map_differencing}. The colors along the bottom correspond to object colors in Figures \ref{fig:jumps_experiment2} and \ref{fig:experiment2_features}.}
  \label{fig:experiment2_images}
\end{figure*}

For basic change detection, we rely on
the \textit{Statistical Inlier Estimation} (SIE) method of
Ekekrantz et al. \cite{ekekrantz2017segmentation}.
The method compares two subsequent local maps from the same location
to find the differences and through them the moving objects. While
the method can also extract parts that are highly dynamic (i.e.
moving while the sweeps are collected), we choose to ignore these parts,
as they mostly consist of moving humans.
SIE explicitly models the noise of the range sensors as part of its optimization,
enabling us to reliably extract smaller objects close by but also
larger objects that are further away.
The resulting detections are fed into the system of Section 
\ref{sec:map_differencing} to add detections of the objects
in the observations where they are static.

In recent years, convolutional neural networks have come to play a
vital role in computer vision. Today, they represent the most 
successful technique e.g. for classification of images.
For applications, a popular method for adopting a network to a particular
domain is \textit{finetuning}. In \cite{chatfield2014return} and \cite{sharif2014cnn}, the authors showed finetuning of networks often
lead to slightly better performance, but can be difficult to get right.
Further, in \cite{sharif2014cnn} as well as \cite{donahue2014decaf},
features extracted from the tensor responses of one of the layers in
a pre-trained network showed remarkable performance on a wide range of computer vision
tasks. Importantly for our application, \cite{sharif2014cnn} showed
these features could achieve state-of-the art performance in both
instance recognition benchmarks and retrieval scenarios.
Whenever we can distinguish the objects by visual appearance we would
like to take advantage of the discriminative power of these features.
Therefore, we use CNN features reduced to three dimensions using
t-SNE \cite{maaten2008visualizing}, see Figure \ref{fig:features2d}.
In our experiments, we found that three dimensions provide a good
trade-off between computational complexity of the filter,
accurate estimation of covariances and discrimination. In
addition, we have seen that after t-SNE reduction, the variation
within each class is well approximated by a multivariate Gaussian.
The two-dimensional illustration in Figure \ref{fig:features2d}
gives some indication of this.

\subsection{Parameters}
\definecolor{lightgray}{gray}{0.9}

\begin{table}[htpb]
\begin{center}
\rowcolors{1}{}{lightgray}
\begin{tabular}{r|rrrrrrr}
 Parameter & $p_{\text{jump}}$ & $p_{\text{meas}}$ & $\sigma_q$ & $\sigma_r$ & $A_k$ & Particles & $\mathbf{R}_k^f$, $S^f$ \\
  \hline
  Value & 0.03 & 0.98 & 0.35 & 0.15 & $20 m^2$ & 300 & estimated \\
\end{tabular}
\end{center}
\label{table:parameters}
\end{table}
Most of the parameters have some meaning and can to an extent be selected by hand.
In these experiments, we have $p_{\text{jump}} = 0.03$, $p_{\text{meas}} = 0.98$,
$\mathbf{Q}_k^s = \sigma_q^2 \mathbf{I}$, $\mathbf{R}_k^s = \sigma_r^2 \mathbf{I}$, with $\sigma_q = 0.35$, $\sigma_r = 0.15$.
Note that the spatial dimensions are independent.
We deliberately set $p_{\text{jump}}$ to a slightly lower value to avoid sampling
too many jumps, which might lead to particle depletion with many objects.
For simplicity, the spatial process noise $\mathbf{Q}_k^s$ is constant,
but could also increase as some function of the time between
observation $k-1$ and $k$.
$\mathbf{Q}_k^f = \mathbf{0}$ since we do not expect the descriptors to change over time.
We fit the measurement covariance $\mathbf{R}_k^f$ and support $S^f$ from the data
set referenced in Figure \ref{fig:features2d}.
While the areas of the locations, $A_k$, can be estimated from
the 3D map observations, the rooms in our data sets are all around
$20 m^2$ and we use this value in all experiments.

\subsection{Evaluation metrics}

To compare the performance of different methods,
we use the CLEAR Multiple Object Tracking metrics
from \cite{bernardin2008evaluating}. The metrics are defined using the
association of targets to observations based on
estimated positions. These associations can be used
to compute the mean errors, number of \textit{mismatching} associations, number of \textit{false positives} and
the number of \textit{missed} observations.
The MOTA metric \cite{bernardin2008evaluating} combines
several of these measures into a global score, showing
the performance of the trackers.
We also compute the MOTP metric, which is simply the
mean distance error of the associations \cite{bernardin2008evaluating}.

To find the best association for the MOTA calculation, we use the
\textit{Hungarian algorithm} to minimize the combined distance between filter estimates
and observations, similar to \cite{bernardin2008evaluating}.
We only consider associations closer than $0.5m$. Once we have computed
the associations, we compute the \textit{false positives} as the number of
estimates associated with a measurement with no label. \textit{Missed observations}
are similarly labels with no estimate. \textit{Mismatches} are observations
which the estimate assigns a label different from the annotated label.
In \cite{bernardin2008evaluating}, the authors count mismatches only in the observation sequence where the initial error is made; for example when two target paths cross.
However, for our application it is important we maintain tracking of
the correct targets over the whole sequence. We therefore count any
observation with the wrong label estimate as a mismatch.

To establish an independent baseline to compare against, we also used
our detections and features in a system similar to GATMO \cite{gallagher2009gatmo}.
Since all short-term dynamics and static structures
have already been filtered, we only
need to keep track of the long-term dynamics.
Therefore, the baseline tracker
keeps track of two sets of objects, \textit{movable} and \textit{absent}.
We use the information from Section \ref{sec:map_differencing} to know
if a detected object was propagated, i.e. if it has not moved.
Given the previous sets of movable and absent objects, a movable
object is explained and kept in the movable set if it is static as
compared to the previous observation. If it is not, or the object was
in the absent set, it can be matched to one of the unexplained
observations.
Then, a match is made if the visual feature distance is smaller than some threshold.
We find the threshold by running several
experiments with different thresholds, and picking the one that
gives the best results on both annotated data sets.
If it is matched in this way, it is placed in the movable set,
otherwise among the absent objects. The estimated position of
a given object is simply taken to be that of its last matched observation.

\subsection{Experimental setup}

We perform several experiments using the
STRANDS robot platform \cite{hawes2016strands}.
In each case, the inputs to the system are RGBD frames a
robot collects autonomously while moving between a few 
different locations.
Typically, after observing one location, the robot visits the other ones
before returning to the same location again.
Since we want to validate the system in the presence of lots of movement,
and especially movement of the objects between different locations,
we manually move the some of the objects when the robot is away.
This gives us ground truth knowledge of which specific objects moved.
As the motion of single objects in the experiments was exaggerated as
compared to most natural environments, we argue that it provides
a thorough validation of the robustness of the tracker.
The collected frames are fused into maps, mostly consisting of an entire room, and
segmented using the system in Section \ref{sec:map_differencing}.
Features are extracted from the images associated with the point cloud
segments. The object position observations are simply computed as the mean
of the point clouds. To initialize the tracking system, we mark the
segments corresponding to the objects we wish to track in the first observations.

The first experiment should verify the tracking ability under frequent movement,
but with dissimilar objects, to avoid ambiguity.
There are a couple of instances of most types, but
in particular the fire extinguisher and monitor types only appear once,
see Figure \ref{fig:experiment2_images}.
In the sequence of 77 observations, the 13 objects jump a total of 14 times,
see Figure \ref{fig:jumps_experiment2}.
While the first experiment should validate the ability to track in
favorable circumstances, the second experiment tests the ability to handle
ambiguity and noise in the features.
To that end, all of the objects are chairs, and most of them are 
visually similar. The seven chairs in the sequence are all in either of two 
locations. In the 55 observations of the locations, there are three jumps 
taking place, as well as considerable local movement of 
all chairs, see Figure \ref{fig:jumps_experiment1}.
The second experiment should be challenging for methods that do
not explicitly handle ambiguous evidence, such as the baseline.

To verify how much information either modality contributes to the tracker,
we benchmark the full system as well as the local and global tracking components
individually. The local tracker is simply a variant of the full system
with $p_{\text{jump}} = 0$, disallowing any jumping within or between locations.
A global tracker based purely on features is obtained by replacing the
spatial Kalman measurement likelihood with a uniform distribution. The spatial position
thus has no contribution to the final estimate. To attain
position estimates, we instead set the positions of each
particle's targets to those of their last estimated measurement correspondences $c_{j, k}$.

Finally, we perform experiments in simulation to investigate how tracking 
performance is affected by varying the number of tracked objects and the rate of jumps.
Given a number of pre-defined areas, jumps are sampled between the areas as well
as Gaussian process noise around the current positions. Additionally, with each
observation round we sample a number of noise detections not corresponding to
any of the targets.
Since we sample a full simulation sequence for every value of the parameters,
the sampling will introduce some additional variability in the
performance measures. To decrease the effects of sampling, we sample three simulations per value, and evaluate the tracker ten times
on every sampled simulation.

\section{Results}

Since the filtering method is sampling based and the results vary slightly, we do 50 experiments with each method and present both qualitative and quantitative results from
these runs. The experiments are all with 300 particles, which proved to
be sufficient to track the posteriors (see Section \ref{sec:various_objects_results}).
Qualitative results are reported on the version using
approximate rejection sampling of the proposal, with
comparisons to the results with other variations
of our method and to the baseline.
We evaluate sampling the proposals using Gibbs sampling
by itself and then also with Gibbs weight calculation.

\subsection{Experiment 1: Various Objects}
\label{sec:various_objects_results}
\begin{figure}[thpb!]
	\centering
 	\includegraphics[trim=45 0 70 30,clip,width=0.99\linewidth]{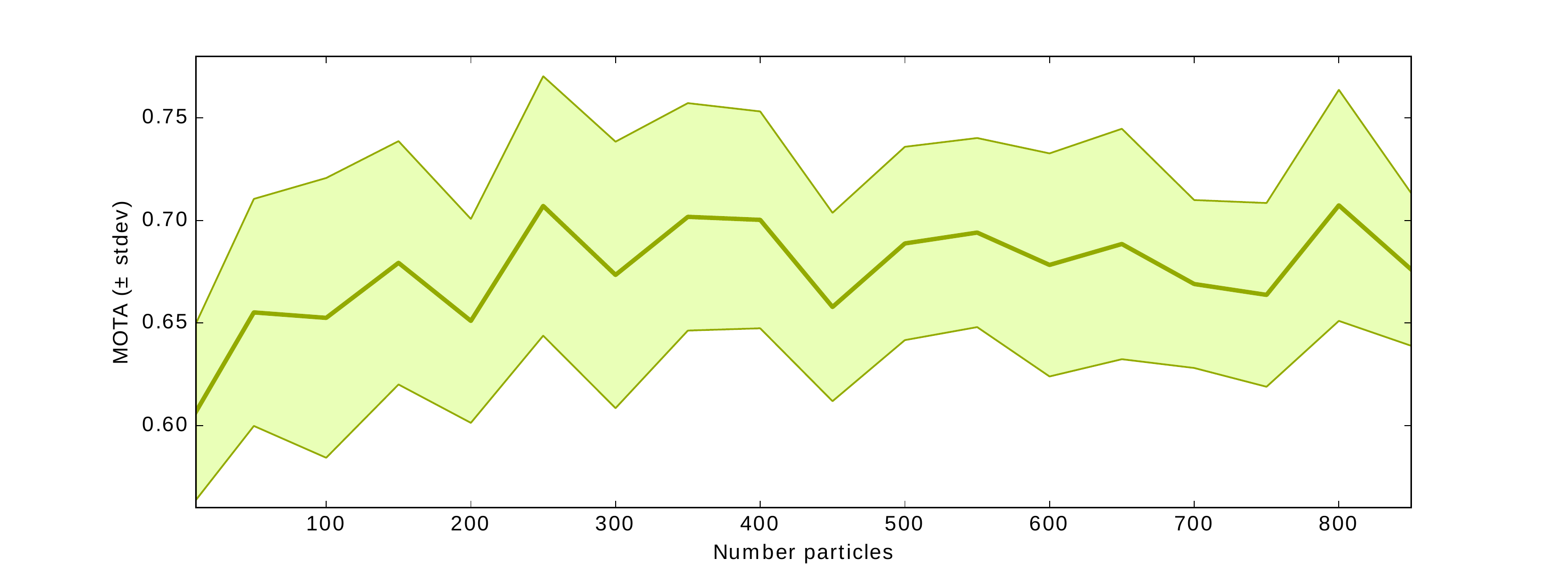} 
    \caption{The MOTA score as a function of the number of particles, with statistics from 10 runs on experiment 2. After about 300, increasing the number of particles has little effect.}
    \label{fig:number_particles_trend}
\end{figure}

In Figure \ref{fig:number_particles_trend}, we see that the MOTA
score on this experiment increases with the number of particles.
After about 300 particles, the trend stabilizes at around $0.7$,
demonstrating that 300 is enough to approximate our
defined posterior accurately, and that other errors are due to
modeling imperfections or system noise.
The posteriors of four different objects from experiment 1 are visualized in Table
\ref{table:posteriors}. Whenever an object disappears from a location,
it affects the posterior of the corresponding object. The local
uncertainty increases, at the same time as new associations are sampled
around other measurement locations. In the example posteriors, we
see that the posteriors are well concentrated when the objects are static.

Features of the tracked objects in experiment 1 are shown in Figure \ref{fig:experiment2_features}. The different object classes
result in features that are well separated. Out of the
ten experiments, 10 out of the 14 jumps are tracked correctly in the
majority of cases. These include successful tracking of diverse objects
such as mugs, bottles, canisters, trash cans and bags.
In fact, the failures were all with larger objects such as the jumping
chair and the monitor. In these cases, the features seem to have
dramatically changed before or during the jumps of the objects. In the
case of the office chair, a rotation led to the feature of another chair
to become more similar, causing confusion in the inference. The other failure occurred at time steps 37 and 39 when the
fire extinguisher and a trash can respectively switched places, see Table
\ref{table:posteriors} and Figure \ref{fig:jumps_experiment2}.
This typically lead the tracker to estimate that the objects stayed in the same places and adapt to the new appearance.
Besides the ten successfully tracked jumps, the two static objects were
inferred correctly.

\definecolor{lightgray}{gray}{0.9}

\begin{table}[htpb]
\begin{center}
\rowcolors{1}{}{lightgray}
\begin{tabular}{r|rrrrr}
 System & MOTP\cite{bernardin2008evaluating} & Miss rate & False pos. & Mism. & MOTA\cite{bernardin2008evaluating} \\
  \hline
  Simplified & 0.11 & 0.17 & 0.07 & 0.08 & 0.67 \\
  Gibbs sampl. & 0.12 & 0.17 & 0.06 & 0.07 & 0.70 \\
  Gibbs weights & 0.12 & 0.17 & 0.04 & 0.06 & 0.73 \\
  Gibbs weights\textsuperscript{*} & 0.11 & 0.13 & 0.01 & 0.00 & 0.86 \\
  Features\textsuperscript{**} & 0.18 & 0.40 & 0.07 & 0.06 & 0.46 \\
  Local\textsuperscript{***} & 0.13 & 0.21 & 0.04 & 0.27 & 0.48 \\
  Baseline & 0.11 & 0.08 & 0.01 & 0.00 & \bf{0.91} \\
\end{tabular}
\end{center}
\caption{Results from experiment 1.
The main method is run with the simplified independent proposal,
Gibbs sampling of proposals and Gibbs estimation of weights.
The rates are averaged over 50 runs. \footnotesize{\textsuperscript{*}optimized for experiment ($0.3 \times \mathbf{R}_k^f$) \textsuperscript{**}no spatial tracking \textsuperscript{***}$p_{\text{jump}} = 0$}}
\label{table:experiment2}
\end{table}

\definecolor{color_3}{rgb}{0.33,  0.22,  0.48}
\definecolor{color_5}{rgb}{1.0, 0.70, 0.0}
\definecolor{color_6}{rgb}{0.58,  0.67,  0.0}
\definecolor{color_8}{rgb}{0.50,  0.24,  0.46}

\newcolumntype{C}{>{\centering\arraybackslash}m{11.5em}}
\begin{table*}\sffamily
\centering
\begin{tabular}{l*5{C}@{}}
\toprule
 & Initialization & 1 & 2 & 3 & 4 \\ 
\midrule
\cellcolor{color_3!25}\rotatebox[origin=c]{90}{Object 3 - Bottle} & \includegraphics[width=0.99\linewidth]{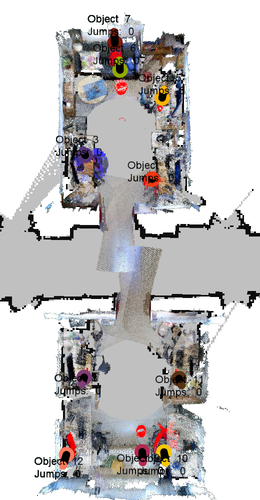}\llap{\includegraphics[height=2cm]{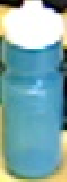}} & \includegraphics[width=0.99\linewidth]{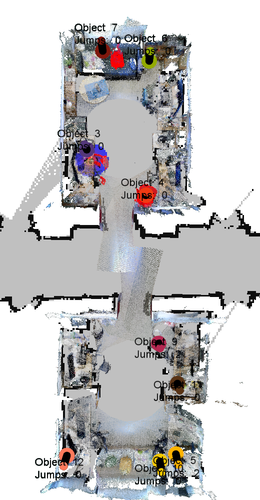}\llap{\includegraphics[height=2cm]{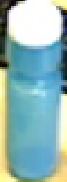}} & \begin{overpic}[width=0.99\linewidth]{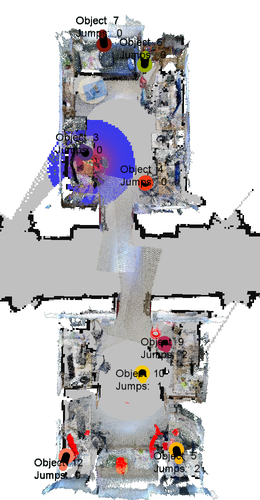}
  \put (12,43) {\huge Jump}
\end{overpic}\llap{\includegraphics[height=2cm]{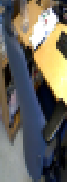}} & \includegraphics[width=0.99\linewidth]{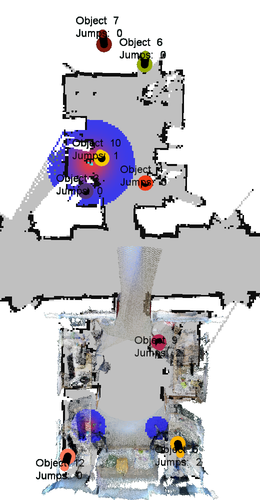} & \includegraphics[width=0.99\linewidth]{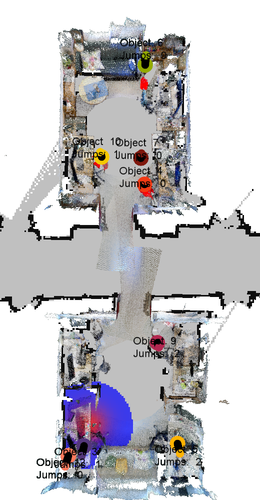}\llap{\includegraphics[height=2cm]{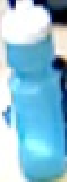}} \\ 
\cellcolor{color_5!25}\rotatebox[origin=c]{90}{Object 5 - Mug} & \includegraphics[width=0.99\linewidth]{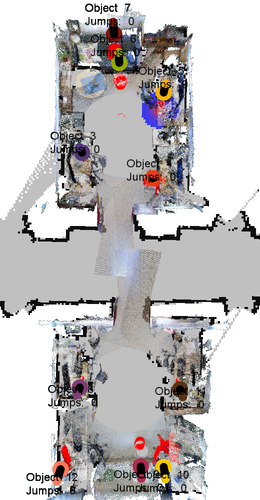}\llap{\makebox[0.99\linewidth][l]{\includegraphics[height=1.5cm]{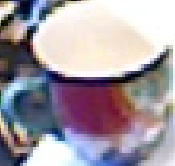}}} & \begin{overpic}[width=0.99\linewidth]{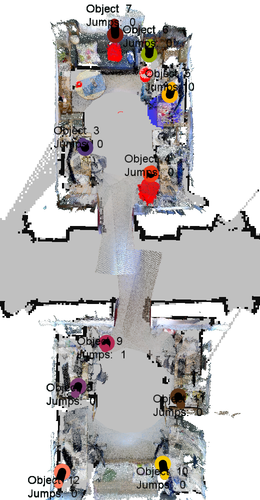}
  \put (12,43) {\huge Jump}
\end{overpic}\llap{\makebox[0.99\linewidth][l]{\includegraphics[height=1.5cm]{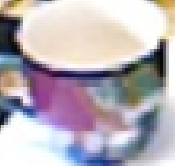}}} & \includegraphics[width=0.99\linewidth]{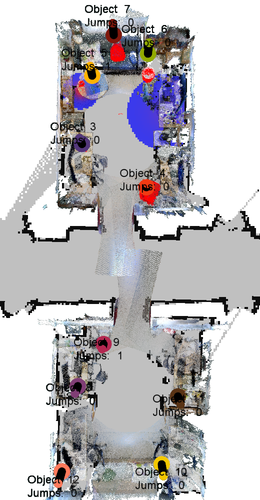}\llap{\makebox[0.99\linewidth][l]{\includegraphics[height=1.5cm]{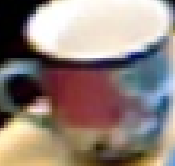}}} & \begin{overpic}[width=0.99\linewidth]{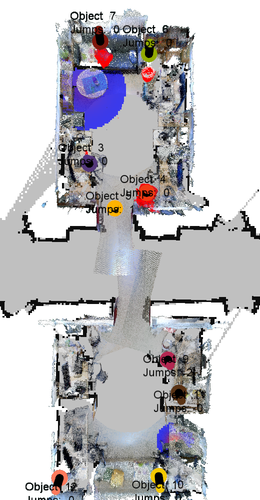}
  \put (12,43) {\huge Jump}
\end{overpic} & \includegraphics[width=0.99\linewidth]{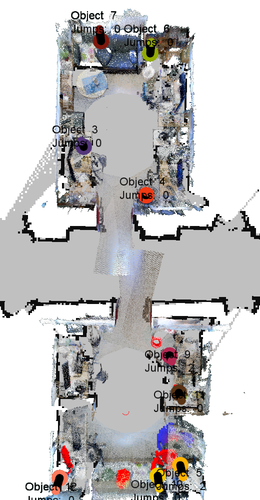}\llap{\makebox[0.99\linewidth][l]{\includegraphics[height=1.5cm]{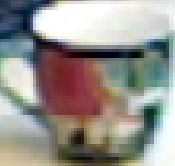}}} \\
\cellcolor{color_6!25}\rotatebox[origin=c]{90}{Object 6 - Trash can} & \includegraphics[width=0.99\linewidth]{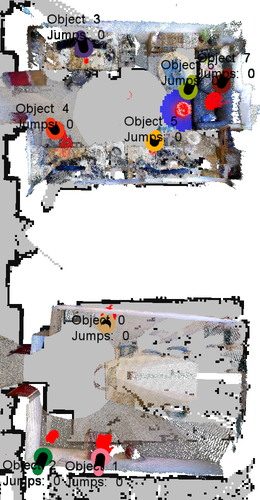}\llap{\includegraphics[height=2cm]{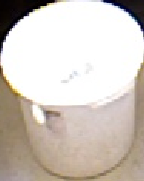}} & \includegraphics[width=0.99\linewidth]{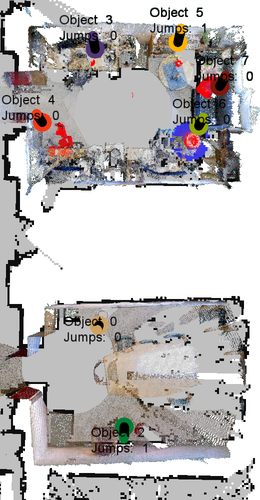}\llap{\includegraphics[height=2cm]{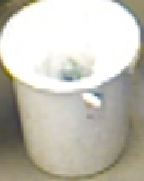}} & \begin{overpic}[width=0.99\linewidth]{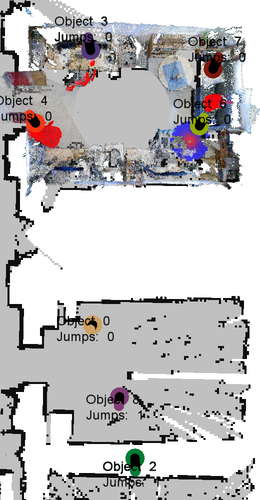}
  \put (12,43) {\huge Jump}
\end{overpic} & \includegraphics[width=0.99\linewidth]{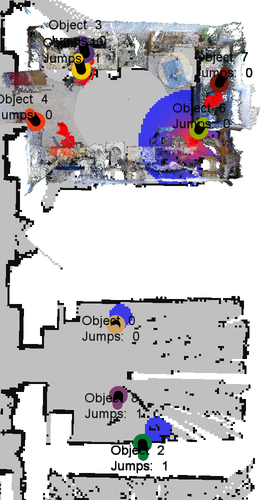}\llap{\includegraphics[height=2cm]{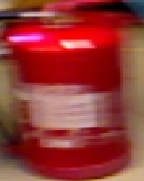}} & \includegraphics[width=0.99\linewidth]{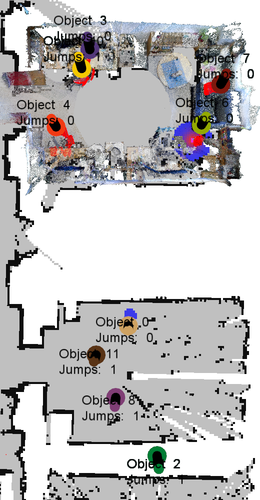}\llap{\includegraphics[height=2cm]{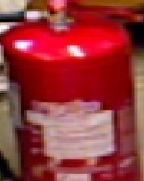}} \\
\cellcolor{color_8!25}\rotatebox[origin=c]{90}{Object 8 - Container} & \includegraphics[width=0.99\linewidth]{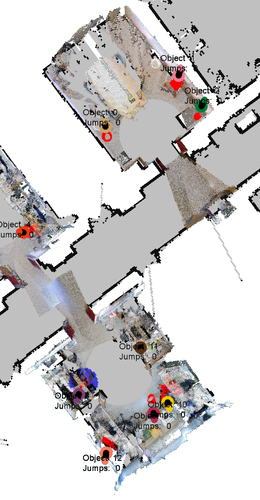}\llap{\includegraphics[height=2cm]{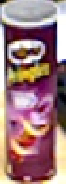}} & \includegraphics[width=0.99\linewidth]{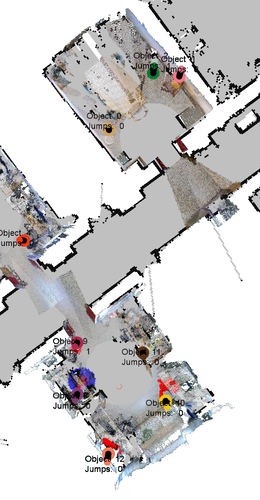}\llap{\includegraphics[height=2cm]{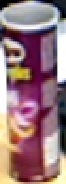}} & \includegraphics[width=0.99\linewidth]{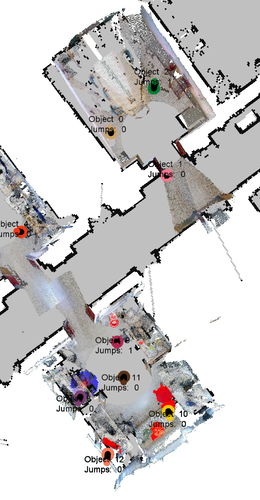}\llap{\includegraphics[height=2cm]{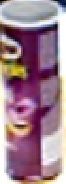}} & \begin{overpic}[width=0.99\linewidth]{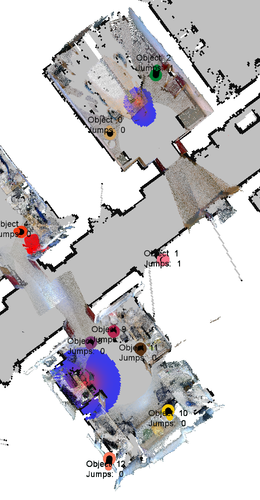}
  \put (12,43) {\huge Jump}
\end{overpic} & \includegraphics[width=0.99\linewidth]{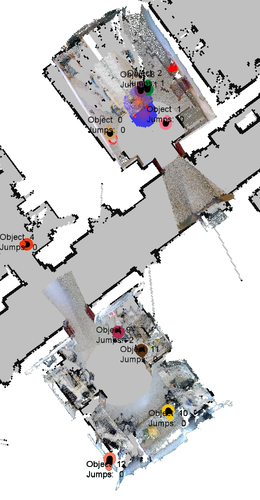}\llap{\includegraphics[height=2cm]{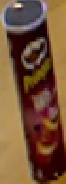}} \\ 
\bottomrule 
\end{tabular}
\caption{Posteriors (red-blue gradients) of four different objects from experiment 1 at different time steps together with the image of the closest measurement.
If there is no image, no measurement was associated with the estimate at that time step. The trash can in column 3 shows one of the failures. Since that and the fire extinguisher swap places around step 2, the estimates are confused, causing a mismatch. The other objects are correctly tracked, since the images of "Intialization" and "4" coincide.
Note that all objects jump at least once.}
\label{table:posteriors}
\end{table*}

Quantitative results are presented in Table \ref{table:experiment2}.
Importantly, we observed that in the experiments where the results
were qualitatively worse, this correlated with a significantly lower MOTA rate.
As expected, with almost no overlap between the feature distributions,
the baseline achieves good results. Compared to the baseline, the proposed
tracker takes several time steps to converge on the new target location
whenever a target jumps. This is due to the uncertainty embedded in the
tracker, and leads to a significantly lower MOTA score in this case.
If we decrease the feature uncertainty to about one third of the estimated covariance $\mathbf{R}_k^f$,
as shown in Table \ref{table:experiment2}, the proposed tracker becomes more certain
and collapses on new locations more quickly, leading to a score almost
in line with the baseline. In general, the proposed tracker infers
the same jumps as the baseline, but estimates are lagging due to uncertainty.

\begin{figure*}[thpb!]
\centering
\begin{minipage}[t]{.49\linewidth}
 	\includegraphics[trim=45 0 70 20,clip,width=0.99\linewidth]{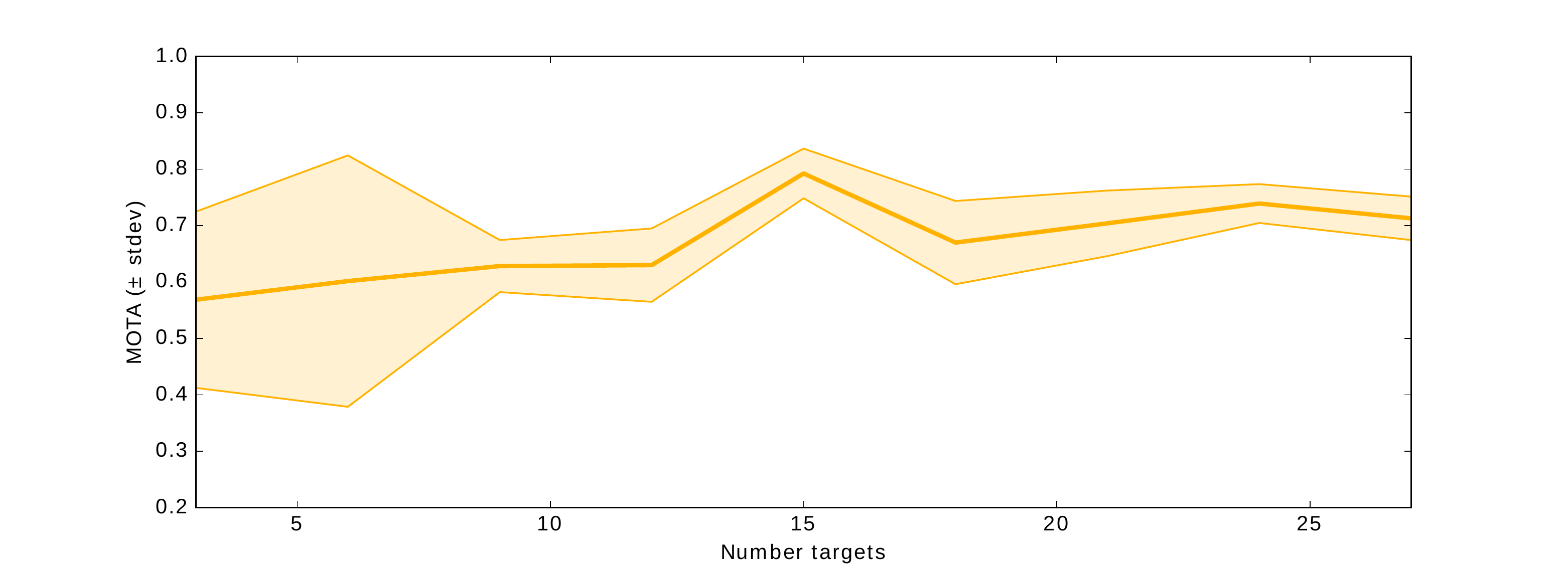} 
    \caption{The MOTA score as a function of the number of targets in simulation. Note that the score stays constant when adding targets.
    The initial variability is due to how the simulation is set up; 
    if only one target is at a location and it jumps somewhere else,
    the old location is unlikely to be observed again, leading the
    tracker to believe that the target is still there.}
    \label{fig:number_targets_trend}
\end{minipage}\hfill
\begin{minipage}[t]{.49\linewidth}
 	\includegraphics[trim=45 0 70 20,clip,width=0.99\linewidth]{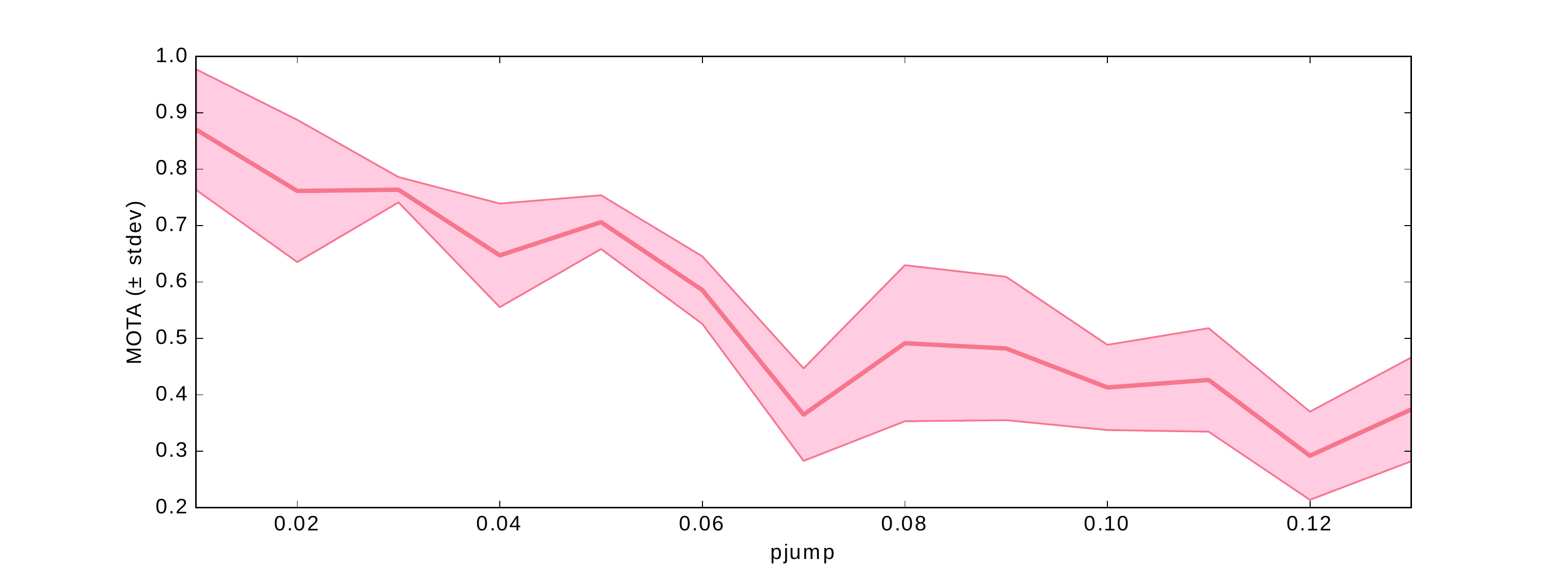} 
    \caption{MOTA score as a function of the rate of jump in simulation, with statistics from 30 runs. Note that the score is decreasing with the jump rate. Since  three simulations are sampled for each rate, some additional variability can be seen.}
    \label{fig:pjump_trend}
\end{minipage}
\end{figure*}

From the results, we see that the local tracker performed
significantly worse than the full system on this dataset.
Due to the large number of jumps, the model is too limited in this scenario. 
The high mismatch rate of the local tracker is due to its inability
to jump, instead settling for the neighboring observations which might be
visually dissimilar.
On the other hand, we see that the feature-based tracker was almost on par
with the local model. However, the high estimated uncertainty of the
features keeps the feature-based tracker from achieving a result in line
with that of the baseline. The result tells us approximately how
much of the performance of the combined model is purely due to the features.
We can see that, with the feature uncertainty, the combined model
performs significantly better than the individual models.
This is mirrored in the qualitative results, where
the full inference framework clearly outperformed both methods.

In this experiment, each Gibbs sampling method contributes
to a slight performance improvement over rejection sampling.
Qualitatively, the Gibbs strategies were able to more often 
correctly infer that the garbage can and fire extinguisher switched 
places and, more generally, collapse on the correct locations faster.
These results and those from experiment 2 give us confidence that while
the MCMC schemes are slightly more accurate, the approximate version is a
reasonable approximation.
The trade-off becomes apparent in the run-time of the
method, which is roughly doubled when doing 100 iterations of burn in for Gibbs sampling.

\subsection{Experiment 2: Chairs}

As can be seen in Figure \ref{fig:experiment1_features},
the features do not discriminate well between the chairs and the inference therefore has
to rely more on the measurement positions to estimate the target positions.
Of the three jumps, the filter tracks two of them correctly,
while the last one happens too late in the observation sequence
for the estimate to consolidate on the new position.
In general, the filter requires two to three observations before a jump is inferred.
At the first jump, the chair jumps close to another identical chair, leading
the tracker to confuse the two chairs in the majority of the ten runs.
Such errors are to be expected as there is no way to distinguish
these visually identical chairs.
Chair number 4 (see Figure \ref{fig:jumps_experiment1})
moves around within a roughly three meter diameter
but the tracker accurately follows it in most of the runs.
The rest of the objects all move around locally and are tracked correctly in
the majority of experiments.

\definecolor{lightgray}{gray}{0.9}

\begin{table}[htpb]
\begin{center}
\rowcolors{1}{}{lightgray}
\begin{tabular}{r|rrrrr}
 System & MOTP\cite{bernardin2008evaluating} & Miss rate & False pos. & Mism. & MOTA\cite{bernardin2008evaluating} \\
  \hline
  Simplified & 0.18 & 0.18 & 0.03 & 0.11 & 0.68 \\
  Gibbs sampl. & 0.18 & 0.15 & 0.02 & 0.12 & \bf{0.70} \\
  Gibbs weights & 0.18 & 0.19 & 0.04 & 0.09 & 0.68 \\
  Features\textsuperscript{*} & 0.25 & 0.84 & 0.01 & 0.04 & 0.11 \\
  Local\textsuperscript{**} & 0.18 & 0.17 & 0.04 & 0.12 & 0.67 \\
  Baseline & 0.16 & 0.20 & 0.02 & 0.28 & 0.49 \\
\end{tabular}
\end{center}
\caption{Results from experiment 2 with chairs. The rates are an average over 50 runs. \footnotesize{\textsuperscript{*}no spatial tracking \textsuperscript{**}$p_{\text{jump}} = 0$} }
\label{table:experiment1}
\end{table}

Quantitative results are presented in Table \ref{table:experiment1}.
Both Gibbs proposal sampling and with weight calculation
have about the same MOTA score as approximate rejection sampling,
with pure Gibbs proposal sampling being slightly higher.
Correspondingly, qualitative results for all three methods are similar.
The baseline method performs significantly worse than these
methods on this experiment. It suffers since it only tracks
static objects spatially, without a soft local movement prior.
Whenever there is any movement, it therefore relies on the
uncertain features to estimate associations. Qualitative
results reflected the low score, with the estimated positions
frequently jumping between different positions.

Since there are few jumps in this experiment, we see that the local
version of the tracker performs well also in comparison to the full system.
In particular, since the jumps do not occur until halfway through the
sequence, this version typically gives perfect results up until that point.
However, the larger jumps cannot be tracked, and it errs at this point.
Interestingly, the small jump within one room
at the end of the sequence is typically inferred correctly.
The feature-based tracker qualitatively performed badly for this experiment,
which can also be seen from the MOTA. Since different particle hypotheses
were typically associated with several of the chairs in the two rooms,
the estimate ended up somewhere in the middle. This can be seen by the
high miss rate, which is due to the position annotations being too far
away from any estimates.
In conclusion, we see the combined tracker
performed the best in this scenario, with the local tracker providing
most information for the joint inference.

\subsection{Experiment 3: Simulations}

In the simulated environment, we generated results for various
number of targets and rate of jumps, with three simulations sampled
for every parameter value. Our hypothesis is that tracking performance
should decrease with the rate of jump, rather than with the number
of targets. This is due to how our proposal distribution is
constructed; when one target is well explained by a measurement,
most of the particles are likely to sample that association.
Hence, particle depletion happens more rapidly when there are
several uncertain associations, as happens when targets jump.
If we look at the graph in Figure \ref{fig:pjump_trend}, we can
indeed see that the MOTA develops negatively with an increasing
jump rate. Correspondingly, Figure \ref{fig:number_targets_trend}
shows the performance stays more or less constant when
increasing the number of targets. From these graphs, we can
deduce that the complexity of the proposed algorithm scales
with the jump rate rather than with the number of targets.
This is important in real world environments, where the number
of objects might be very large, but jumps happen relatively seldom.

\subsection{Data from robot deployment}
\label{sec:real_data}
\definecolor{lightgray}{gray}{0.9}

\begin{table}[htpb!]
\begin{center}
\rowcolors{1}{}{lightgray}
\begin{tabular}{r|rrrrr}
 System & MOTP\cite{bernardin2008evaluating} & Miss rate & False pos. & Mism. & MOTA\cite{bernardin2008evaluating} \\
  \hline
  Proposed & 0.14 & 0.15 & 0.10 & 0.04 & 0.71 \\
\end{tabular}
\end{center}
\caption{Statistics from robot deployment, with 9 objects
         in 34 observations. Values are comparable to other experiments.}
\label{table:real_experiment}
\end{table}
In addition to the experiments, we have also evaluated the method
on data gathered by a mobile robot in a real world deployment.
The robot patrolled an office environment during a period of
two months, gathering observations of a few places with one
or two days' interval. In our analysis, we have focused on
three places with highly movable objects: one reception
area and two kitchens. In these scenes, we marked three objects
in each location, most of them moving around within the area.
Since no motion was observed in between the locations, the experiment
should mostly be seen as a way to test the stability of
the methods in a real world scenario.
In Table \ref{table:real_experiment}, we see that the numbers
reflect those in other experiments. Indeed, the method manages
to track the objects through most of the sequences, confirming
the validity of the approach as applied to natural data.

\section{Discussion}

We see that the joint tracker is able to bring in information from both
the visual and spatial modalities to produce a reliable estimate of the
object positions. In the case where we have different object classes,
the CNN features proved to be highly discriminative,
allowing us to track the
jumps even of comparatively small cups. In fact, the small objects seem to
consistently provide good features, even in the presence of motion blur.
This, together with the preliminary deployment results in Section \ref{sec:real_data},
gives us confidence that the proposed system will be able to work in
a real world scenario, where change detection is likely to detect the movement
of a multitude of small objects.
Further, the method works well even when we increase the
number of objects to track since the accuracy of the
algorithm depends more on the number of jumps.
To maintain accurate posteriors in the event of many jumping objects,
one could employ an adaptive particle set scheme such as \cite{fox2003adapting}.

The baseline method is based on the previous GATMO \cite{gallagher2009gatmo},
as applicable within our system. Since this method does not
reason probabilistically, it suffers when there is noise and ambiguity
in the measurements, as can be seen in the second experiment.
Since there is overlap between the distributions of the different
features, objects with overlap are sometimes confused.
In a real-world scenario, this kind of situation is to be expected,
as there are often many similar-looking objects. Moreover, we
expect more noise in real-world scenarios, when there is more lighting
variation, and the objects may be observed from more angles.
We conclude that it is important to incorporate uncertainty as part of the tracking,
especially in environments that contain multiple visually similar objects.

While not discussed in the results, the tracking is enabled partly by
the performance of the change detection framework discussed in Section
\ref{sec:map_differencing}. Even when the objects did not move, and were
detected through the propagation approach, all objects were detected in all
the observations. This is remarkable, since in total, we have
522 annotated objects in the datasets.
It shows that this simple method is sufficient, at least given precise registration.
However, the segmentation sometimes
either over- or undersegments the objects. Particularly for the chairs,
oversegmentation is an issue that should be dealt with in a more principled
fashion, potentially by training a supervised segmentation method on similar data.
In most cases, the tracker was nonetheless able to handle these deficiencies.

\begin{figure}[thpb!]
	\centering
 	\includegraphics[trim=20 70 20 80,clip,width=0.99\linewidth]{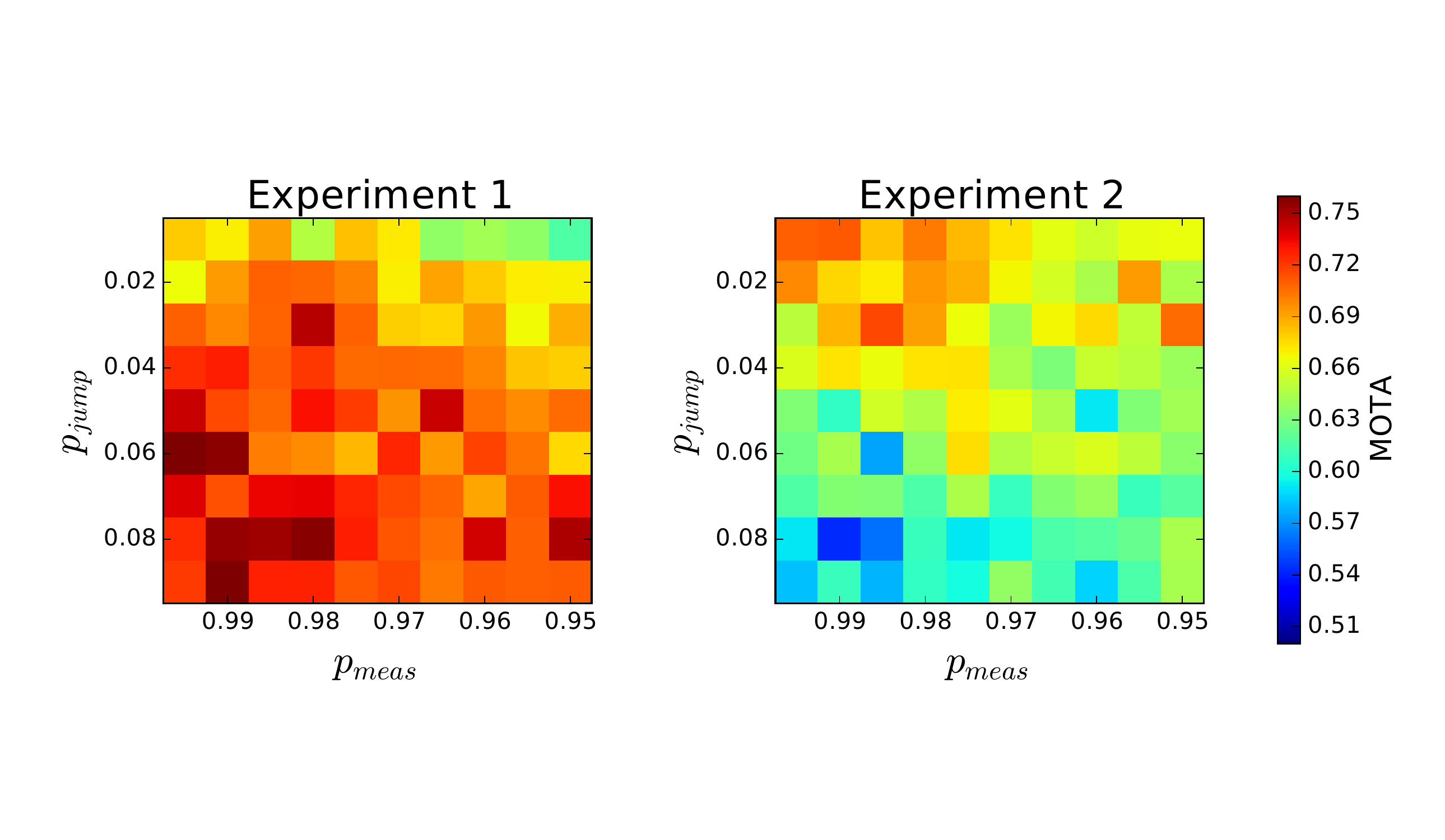} 
    \caption{Effect of the interaction of $p_{\text{jump}}$ and $p_{\text{meas}}$ on the MOTA score in the two experiments. In experiment 2, the method performs well roughly within the quadrant $p_{\text{jump}} \leq 0.04$, $p_{\text{meas}} \geq 0.975$. The 
    optimal value of $p_{\text{jump}}$ seems to be higher in the first experiment, at or above $0.08$.}
    \label{fig:pjump_pmeas}
\end{figure}

In Figure \ref{fig:pjump_pmeas}, we see the effect of varying
the $p_{\text{jump}}$ and $p_{\text{meas}}$ parameters. In the
experiments, we have chosen to use the same values of the
parameters for consistency. However, from the figure, we observe
that the method performs better on experiment 1 when the value
is higher than the one used. The difference in the optimal value
of $p_{\text{jump}}$ between the experiments is due to that the
actual rate of jumps is higher in experiment 1, with about
$0.18$ jumps per observation versus $0.05$ in experiment 2.
The relation in the jump frequencies to the optimal $p_{\text{jump}}$
values leads us to believe that it might be beneficial to estimate the
number of jumps using the algorithm and adjust the
$p_{\text{jump}}$ parameter iteratively in an EM fashion.
Similarly, we saw in experiment 1 that estimating the feature covariance
$R^f$ for the data at hand can significantly improve results.

\section{Conclusion \& Future Work}

We presented a system which can track multiple similar objects even in the 
presence of noise and when only a subset of the objects can be
observed at any given time. There are three major reasons that together enable the
system to track objects of sizes that range from mugs to chairs accurately.
First, the probabilistic method of \cite{ekekrantz2017segmentation} together with our temporal segmentation
allows us to identify even small changes in the 3D maps consistently.
Secondly, recent CNN architectures such as the one used \cite{szegedy2016rethinking} allow us to represent the discovered objects
using discriminative features. For the smaller objects, it is key to use
the image data as the depth data is often too noisy to produce reliable
features. Third, our Rao-Blackwellized object tracker accurately
models the dynamics of the scenario and allows us to maintain realistic
posteriors of object positions. Even when the objects are visually similar,
the tracker can rely on the spatial measurements for reasoning.

As mentioned, one pitfall of the current system is that it relies on a closed world assumption.
Without this, tracking the jumps of objects would likely be intractable.
An open question is how to resolve this conflict. We would like to investigate if we could
model how probable different object types are to stay within the confines of the environment.
For example, a chair is very likely to stay while a cardboard box will likely be thrown away.
It would therefore be appealing to learn these probabilities and track jumps only when the
object is likely to stay in the environment.
For temporary objects, the system should instead track the births and deaths,
see e.g. \cite{oh2009markov}\cite{sarkka2007rao}.
With the ability to integrate new objects as they are observed
comes also the possibility of applying the movement models
to improve a full-blown SLAM system as suggested in \cite{bowman2017probabilistic}.

More generally, we would like to learn more properties of the objects given the visual
features. In addition to estimating the jump probability of the object type, there should
also be specific movement model variances for different targets. In this case, a monitor
serves as a good example of something that stays in the exact same place for a long time,
while a chair will always be expected to move around a bit. The chair should thus
have a higher uncertainty $\mathbf{Q}_k^s$ associated with its movement model
than the monitor.
In turn, one specific kitchen chair might be more
probable to move than an office chair.

\section{Acknowledgements}

The authors would like to thank Erik Ward for many rewarding
conversations on the problems of multi-target tracking.
The work presented in this paper has been funded by the European
Union Seventh Framework Programme (FP7/2007-2013) under grant
agreement No 600623 (``STRANDS'').

\bibliography{tic}
\bibliographystyle{ieeetr}

\copyrightnotice

\end{document}